\pdfoutput=1

\documentclass[11pt]{article}

\usepackage{authblk}

\usepackage[]{ACL2023}

\usepackage{times}
\usepackage{latexsym}

\usepackage{tipa}

\usepackage{enumitem}
\setitemize{noitemsep,topsep=0pt,parsep=0pt,partopsep=0pt}

\usepackage[T1]{fontenc}

\usepackage[T2A,T1]{fontenc}
\usepackage[utf8]{inputenc}
\usepackage[russian,english]{babel}

\usepackage{microtype}

\usepackage{amsmath}
\usepackage{inconsolata}
\usepackage{amsfonts}
\usepackage{multirow}
\usepackage{listings}
\usepackage[ruled,linesnumbered]{algorithm2e}  
\usepackage{CJKutf8}
\usepackage{graphicx}
\usepackage{subfigure}
\usepackage{supertabular}
\usepackage{subfigure}
\usepackage{dcolumn,booktabs}
\newcolumntype{d}[1]{D{.}{.}{#1}}
\newcommand\mc[1]{\multicolumn{1}{c}{#1}} 
\long\def\devour#1{}
\long\def\eat#1{}
\DeclareMathOperator*{\argmax}{arg\,max}

\title{A Crosslingual Investigation of Conceptualization in 1335 Languages}

\author[$\,$]{$\text{\bf Yihong Liu}^{\text{*}\diamond}$,
        $\text{\bf Haotian Ye}^{\text{*}\diamond}$,
        $\text{\bf Leonie Weissweiler}^{\text{*}\diamond}$,
        $\text{\bf Philipp Wicke}^{\text{*}\diamond}$\\
        $\text{\bf Renhao Pei}^{\text{*}}$,
        $\text{\bf Robert Zangenfeind}^{\text{*}}$,
        $\text{\bf Hinrich Sch\"utze}^{\text{*}\diamond}$
}

\affil[*]{Center for Information and Language Processing, LMU Munich} \affil[$\diamond$]{Munich Center for Machine Learning (MCML)
 \protect\\ \texttt{\{yihong, yehao, weissweiler, pwicke\}@cis.lmu.de}}

\def\methodname{Conceptualizer\xspace}

\def\figref#1{Figure~\ref{fig:#1}}
\def\figlabel#1{\label{fig:#1}\label{p:#1}}

\def\tabref#1{Table~\ref{tab:#1}}

\def\tablabel#1{\label{tab:#1}\label{p:#1}}

\def\secref#1{\S\ref{sec:#1}}
\def\seclabel#1{\label{sec:#1}}

\newcounter{notecounter}

\newcommand{\enoteson}{\long\gdef\enote##1##2{{
\stepcounter{notecounter}
{\large\bf
\hspace{1cm}\arabic{notecounter} $<<<$ ##1: ##2
$>>>$\hspace{1cm}}}}}
\enoteson

\makeatletter
\newcommand{\removelatexerror}{\let\@latex@error\@gobble}
\makeatother

\def\mathindent{\mbox{\hspace{0.5cm}}}

\def\concept#1{`#1'}
\def\normalstring#1{``#1''}
\def\algorithmstring#1{$\texttt{#1}$}
\def\meanings#1{\textit{#1}}

\definecolor{atla_color}{HTML}{F51893}
\definecolor{aust_color}{HTML}{005BE0}
\definecolor{indo_color}{HTML}{00A300}
\definecolor{guin_color}{HTML}{F5D000}
\definecolor{otom_color}{HTML}{F56600}
\definecolor{sino_color}{HTML}{A30021}

\definecolor{afr_color}{HTML}{F51893}
\definecolor{eur_color}{HTML}{005BE0}
\definecolor{pap_color}{HTML}{00A300}
\definecolor{nam_color}{HTML}{F5D000}
\definecolor{sam_color}{HTML}{F56600}
\definecolor{aus_color}{HTML}{A30021}

\begin{document}


\maketitle
\begin{abstract}
Languages differ in how they divide up the world into
concepts and words; e.g., in contrast to English, Swahili has a single
concept for \concept{belly} and \concept{womb}. We investigate these
differences in conceptualization across 1,335 languages by
aligning concepts in a parallel corpus. To this end, we
propose \methodname, a method that creates a bipartite
directed alignment graph between source language concepts and sets of
target language strings.  In a detailed linguistic analysis
across all languages for one concept (`bird') and an
evaluation on gold standard data for 32 Swadesh concepts, we
show that \methodname has good alignment accuracy.
We demonstrate the potential of
research on conceptualization in NLP with two experiments. (1) We
define crosslingual stability of a concept as the degree to which it has
1-1 correspondences across languages, and show that
concreteness predicts  stability. (2) We
represent each language by its conceptualization pattern for
83 concepts, and define a similarity measure on these
representations.  The resulting measure
for the conceptual similarity between two languages  is
complementary to standard genealogical, typological, and surface
similarity measures.
For four out of six language families, we can assign languages to their correct family based on conceptual similarity with accuracies between
54\% and 87\%.\footnote{We release our code at \url{ https://github.com/yihongL1U/conceptualizer}}
\end{abstract}



\section{Introduction}
Languages differ in how they divide up the world into
concepts and words. The Swahili word `tumbo' unites the meanings of
the English words `belly' and `womb'. Therefore, English forces its
speakers to differentiate between the general body region
``front part of the human trunk below the ribs'' and one
particular organ within it (the womb) whereas Swahili does
not.  Similarly, Yoruba `irun' refers to both hair
and wool. Again, English speakers must make a distinction
whereas Yoruba has a single hair concept that includes
the meaning \meanings{animal hair for clothing}.

\begin{figure}
  \centering
  \includegraphics[width=0.48\textwidth]{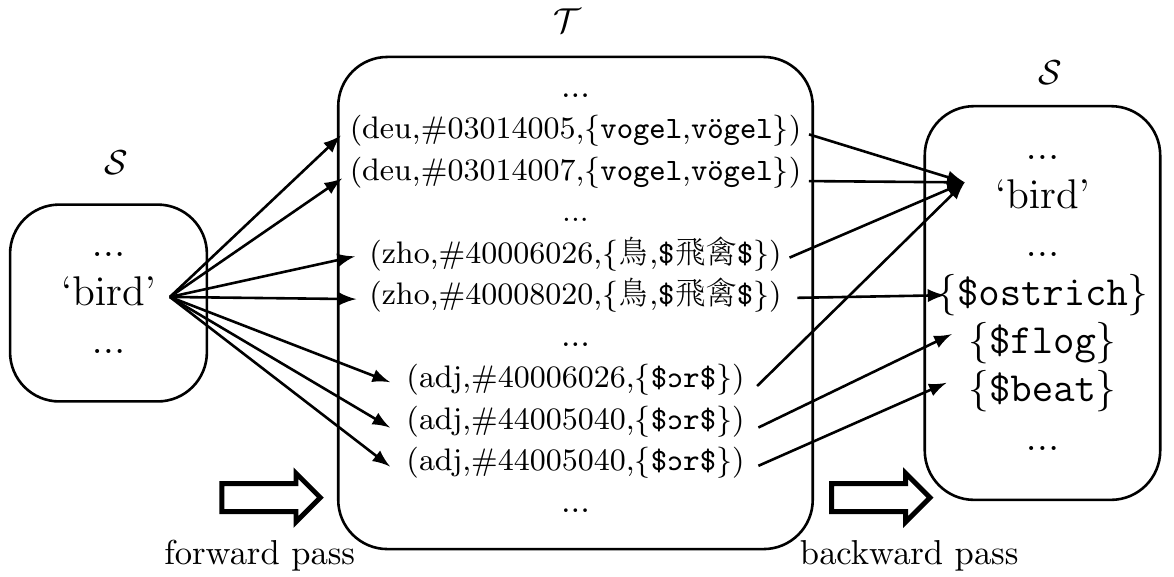}
  \caption{An example of the directed bipartite graph we construct,
  for the concept `bird'. Each node in  $\cal S$ is a set of
  strings. Each node in $\cal T$ is a triple of language,
  verse identifier (i.e., a sentence ID) and set of strings identified
as correlated with `bird'.
\methodname induces edges from both $\cal S$ to $\cal T$ and
  $\cal T$ to $\cal S$ that we then use for analysis and
  prediction. In the example, we see that the set of
  strings correlated in Mandarin (zho) also refers to
  ``ostrich'' and that the correlated string in Adioukrou (adj) is
  ambiguous between ``bird'' and ``flog''.}
  \label{fig:pipeline}
\end{figure}

While studies have looked at
conceptualization within different
languages \cite{ravin2000polysemy,goddard2013words}, we present a
crosslingual study that directly compares conceptualization 
in 1,335 languages. The empirical basis
are word and ngram correspondences in 
the \textbf{P}arallel \textbf{B}ible \textbf{C}orpus
(PBC, \citep{mayer-cysouw-2014-creating}).
We introduce \methodname, a method that
reliably aligns a set of 83 concepts
across all PBC languages.
The 83 concepts are partly chosen to be well represented in
the Bible, and partly from
Swadesh 100 \citep{swadesh2017origin}.
The alignments are formalized
as a bipartite
graph between English (the source) and the target
languages.

The simple idea underlying \methodname
-- illustrated in Figure~\ref{fig:pipeline} --
is that, starting with
one of the 83 concepts in English as the focal concept
(the search query), we can
identify divergent conceptualizations by first searching
for target ngrams highly associated with the focal concept,
and then searching
for English ngrams highly correlated with the
target ngrams we found.  If the English ngrams
correspond to the original focal concept, then the
conceptualizations do not diverge.  In
contrast, take the example of divergence described above: we
start with the focal concept \concept{hair},  find Yoruba `irun'
and then two English concepts, not one, that are highly
associated with `irun': \concept{hair} and \concept{wool}. This indicates that
English and Yoruba conceptualizations diverge
for \concept{hair}.

Our main contribution is that we present the first empirical
study of crosslingual conceptualization that grounds the
semantics of concepts directly in  contexts -- the
sentences of the parallel corpus. This ensures that our work
is based on identical (or at least very similar) meanings
across all 1,335 languages we investigate.
For example, verse
Matthew 9:7 has the same meaning in English:
``Then the man got up and went home.'', in
Chinese:
``\begin{CJK*}{UTF8}{bsmi}那個人就起來，回家去了。\end{CJK*}''
and in each of the
other 1,333 languages.
Such a direct
grounding in meaning across a
large set of languages has not previously been achieved in
work on conceptualization in theoretical or computational linguistics.

In addition, we make the following contributions.
(i) We propose \methodname, an alignment method
specifically designed for concept alignment, that operates on
the level of ngrams and ngram sets.
(ii) We conduct an evaluation of \methodname for the concept
\concept{bird} in all 1,335 languages.
The result is a
broad characterization of how the conceptualization of bird
varies across the languages of the world.
Out of 1,335 languages, \methodname only fails 15
times (due to data
sparseness) for \concept{bird}. 
(iii) We evaluate \methodname
for 32 Swadesh concepts
on a subset of 39 languages
for which translation resources exist and
demonstrate good 
performance.
(iv)
Using the ratings provided by \citet{brysbaert2014concreteness},
we give evidence that 
concreteness (i.e., the degree to which a concept refers to
a perceptible entity)
causes a concept to be more stable across languages:
concrete concepts are more likely to have one-to-one
mappings than abstract concepts.
(v) We propose a new measure of language similarity. 
Since we have aligned
concepts across languages, we can compute measures of how
similar the conceptualization of two languages is. We show
that this gives good results and is complementary to
genealogical, typological and surface similarity measures
that are commonly used.
For example,
Madagascar's Plateau Malagasy is conceptually
similar to
geographically distant typological
relatives like Hawaiian, but also to
typologically distant
``areal neighbors''
like Mwani  and Koti. For four out of six language families,
based on conceptual similarity, we can assign languages to their correct family with between
54\% and 87\% accuracy.


\section{Related Work}
In linguistics, conceptualization has been studied empirically with regards to crosslingual polysemy or colexification \citep{franccois2008semantic,perrin2010polysemous, list-etal-2013-using, jackson2019emotion} as well as areal and cultural influences on concept similarity \citep{gast2018areal,thompson2020cultural, georgakopoulos2022universal}. 
Most of this work is based on human annotations, such as CLICS \citep{list_johann_mattis_2018_1194088, list2018clics2, rzymski2020database}, a database of colexification.  However, the coverage of such resources in terms of concepts included, especially for some low-resource languages, is low. Therefore we explore the use of an unannotated broad-coverage parallel corpus as an alternative.
Expanding this work to many languages is important
to the extent that we accept some (weak) form of linguistic
relativity, i.e., the hypothesis that language
structure (including conceptualization)  influences 
cognition and perception
\citep{boroditsky2003sex,deutscher2010through,goddard2013words}.


Methodologically, our work is closely related
to \citet{Ostling2016colexification} who explores
colexification through PBC. He targets specific
colexification pairs and investigates their geographical
distribution using word alignments. In comparison, our
method allows us to identify alignments beyond the word
level and therefore richer associations among concepts are
obtained. Our proposed method \methodname is also close to
semantic mirrors \citep{Dyvik2004semanticmirrors}, a method
to explore semantic relations using translational
data. The authors focus on an English-Norwegian
lemmatized parallel corpus; in contrast,  we investigate 1,335
languages, most of which are low-resource and for many of which
lemmatization is not available. In addition,
this paper is related to recent work that uses
PBC to investigate the typology of
tense \citep{asgari-schutze-2017-past}, train massive multilingual embeddings \citep{dufter-etal-2018-embedding}, extract multilingual named entities \citep{severini-etal-2022-towards}, find case markers in
a multilingual setting \citep{weissweiler-etal-2022-camel}
and learn language embeddings containing typological
features \citep{ostling2023language}.


Like \methodname, \citet{csenel2017measuring, csenel2018Semantic}
analyzed the semantic similarity of concepts across languages (mainly European ones).
But they use pretrained word embeddings
\citep{word2vec2013mikolov, pennington-etal-2014-glove}, 
which are not available
in high enough quality for most of the low-resource languages we cover in this work.

Computational criteria for language similarity have been
taken from
typology
\citep{ponti,georgi-etal-2010-comparing,pires2019multilingual,daume-iii-2009-non},
morphology \citep{zervanou2014word,
dautriche2017wordform} and language-model surface similarity  \citep{pires2019multilingual, wu2020all}.
We propose a new similarity measure, based on
conceptualization,
with complementary strengths and weaknesses.

There is a large body of work on
statistical and neural word
alignment; recent papers with extensive discussion of this
subfield include
\citep{ho-yvon-2019-neural,zenkel-etal-2020-end,wu-etal-2022-mirroralign}.
We show below that the standard alignment method Eflomal
 \citep{ostling2016efficient} does not work well for our
 problem, i.e., for identifying high-accuracy associations between concepts.

\section{Methodology}
\subsection{Data}
\seclabel{data}
We work with
the Parallel Bible Corpus
(PBC, \citet{mayer-cysouw-2014-creating}).
We use 1,335 Bible translations from PBC, each from a
different language as identified by its ISO 639-3 code.
For most languages, PBC only covers the New Testament (NT)
($\approx$7,900 verses). For a few hundred, it covers both
NT and Hebrew Bible ($\approx$30,000 verses).
See \secref{details_pbc} for details of the PBC corpus.

From Swadesh 100 \citep{swadesh2017origin},
a set of 100 basic universal concepts, we select the 32
concepts that
occur with frequency $5<f\leq 500$ in both NT and Hebrew
Bible.
We call the resulting set of 32 concepts \textbf{Swadesh32}.
We also select
\textbf{Bible51} from the Bible, a set of 51 concepts
that are of interest for crosslingual comparison. Notably,
we include abstract concepts like \concept{faith} that are missing
from Swadesh32.
See \secref{details_concept_select} for concept selection details.

\begin{table}
\centering
{\footnotesize
\begin{tabular}{ll}
&\textbf{Notation}\\\hline
$\cal P$ & power set\\
$\Sigma$ & alphabet\\
$\cal G$ & directed bipartite graph\\
$\cal S$ & the set of source nodes of $\cal G$\\
$\cal T$ & the set of target nodes of $\cal G$\\
$\Lambda$ & the set of 1335 languages\\
$l$ & $l \in \Lambda$, a language\\
$\Pi$ & the set of 31,157 Bible verses\\
$\Pi(l,U)$ & set of  verses of $l$ containing $u \in U$\\
$V$ & $V \subseteq \Pi$, a set of verses\\
$v$ & $v \in \Pi$, a verse\\
$F$ &  focal (English) concept (this is a set), $F \in \cal S$\\
$s$ & source (English) string\\
$S$ & set of source (English) strings\\
$t$ & target string\\
$T$ & set of target strings\\
$U$ & set of  strings (source or target)
\end{tabular}}
\caption{\tablabel{notation}Notation}
\end{table}

\subsection{\methodname}

\textbf{Bipartite graph.} We formalize the concept alignment
graph as a
directed bipartite graph. 
With
$\Sigma$ denoting the alphabet, let
${\cal S} \subset {\cal P}(\Sigma^*)$ be the set of source
nodes, each corresponding to a concept, represented as a set
of strings from the source language; e.g.,
\{\texttt{\$belly\$}, \texttt{\$bellies\$}\} for \concept{belly}, where \texttt{\$} denotes the word boundary.
In this paper, we always use English as the source language.
With $\Lambda$ denoting the set of languages and $\Pi$ the
set of verses,
let
${\cal T} \subset \Lambda \times \Pi \times {\cal P}(\Sigma^*)$ be the set of target
nodes, each corresponding to a
triple of target language $l$, Bible verse $v$ and a set of strings
from language $l$, one of them
occurring in $v$.
We represent
the concept correspondences as a directed bipartite Graph ${\cal G} \subset
{\cal S} \times {\cal T} \cup
{\cal T} \times {\cal S}$, as shown
in \figref{pipeline}. See \secref{additional_details} for
method details.
\tabref{notation} gives our notation.
The reason for our asymmetric design
of the graph (concept \emph{types} on the source side, concept
\emph{tokens} occurring in context on the target side) is that we
want to track how much evidence there is for a
concept-concept correspondence. The more edges there are in the graph, the more reliable the correspondence is.

\begin{figure*}

\begin{tabular}{ll}
\begin{minipage}{7cm}
{\scriptsize
\begingroup
\removelatexerror
 \begin{algorithm}[H]
    \caption{Forward Pass (FP)}  
    \label{alg:fs}  
    \KwIn{focal concept $F$, language  $l$}
    $T \gets \emptyset$\;
\For{$i \gets 1$ \textbf{to} $M$}{
$V\!\gets\!\{ v \!\in\! \Pi |
(\exists s \!\in\! F : s \!\in\! v) \wedge (\neg \exists
t \!\in\! T: t \!\in\! v)\}$\;
\lIf{$i=1$}{$V_1 \gets V$}
            \lIf{$\text{COVERAGE}(l,T,V_1) \geq \alpha$}{break}
        $ t \gets \argmax_{\{t \in {\cal P}(\Sigma^*):
        t \notin T\}} \chi^2(l,t,V)$\;
$T \gets T \cup \{t\}$\;
}
\KwRet{\hspace{5cm}$\{(F,(l,v,T))|\exists t\!\in\! T:   t \!\in\! v \}$}
               \end{algorithm}
               \endgroup}
               \end{minipage}
&
\begin{minipage}{8cm}
{\scriptsize
\begingroup
\removelatexerror
                \begin{algorithm}[H]
    \caption{Backward Pass (BP)}  
    \label{alg:bsnew}  
    \KwIn{focal concept $F$, language  $l$}
$S \gets \emptyset$\;   
\For{$i \gets 1$ \textbf{to} $M$}{
$V\!\gets\!\{ v \!\in\! \Pi|(\exists (F, (l,v,T)) \!\in\! {\cal G})\!\wedge\! (\neg \exists s \!\in\! S: s \!\in\! v)\}$\;
\lIf{$i=1$}{$V_1 \gets V$}
            \lIf{$\text{COVERAGE}(\mbox{eng},S,V_1) \geq \alpha$}{break}
        $ s \gets \argmax_{\{s \in {\cal P}(\Sigma^*) :
        s \notin S\}} \chi^2(\mbox{eng},s,V)$\;
$S \gets S \cup \{s\}$\;
}
\KwRet{$\{((l,v,T), S')|(\exists (F,(l,v,T)) \!\in\! {\cal G}) \!\wedge\! (\exists s \!\in\! S': s \!\in\! S, s \!\in\! v)\}$}
\end{algorithm}
               \endgroup
}
               \end{minipage}
\end{tabular}
\caption{Forward Pass (FP) and Backward Pass (BP) for graph
induction}
\figlabel{fpbp}
\end{figure*}

\textbf{Association for alignment.}
We can represent a source concept (e.g., \{\texttt{\$belly\$}, \texttt{\$bellies\$}\}) as the set of verses $V$ in which
it occurs.
In contrast to standard alignment algorithms, we
exhaustively search all strings $t$ of the target language $l$
for high correlation with $V$.
For example, we search for the French string $t$ that has the
highest correlation with the verses that contain
\{\texttt{\$belly\$}, \texttt{\$bellies\$}\}; the result is
$t$=\normalstring{ventre}. 
This means that we are not limited
to knowing what the relevant (tokenization) units are in advance, which is
not possible for all 1,335 languages.
We use the $\chi^2$ score $\chi^2(l,t,V)$ as a measure of
correlation: we test, for all $t$, whether the two categorical variables
$t \in v$
(short for: $t$ is a substring of verse $v$ in language $l$)
and $v \in V$  are independent. We 
select
the
$t$ with the highest score.

\textbf{Termination.}
For a query string $q$,
e.g., \algorithmstring{\$hair}, occurring in verses $V$,
we want to find a set $U$ of
highly associated ngrams in the target language $l$ that covers
all of $V$.
Because of noise,
translation errors, nonliteral language etc., this is often
impossible. We therefore terminate the
search for additional target strings when
$\mbox{COVERAGE}(l,U,V) \geq \alpha$ where
we set $\alpha = .9$ and
define:
\begin{align*}
    \mathindent\mathindent\mbox{COVERAGE}(l,U,V) = \frac{|\Pi(l,U)\cap V|}{|V|}
\end{align*}
i.e., the fraction of $V$ covered by the strings in $U$.

\textbf{Graph induction.}
\figref{fpbp} shows that
\methodname consists of a
forward pass (FP, Algorithm \ref{alg:fs}) that adds edges $e
\in {\cal S} \times {\cal T}$ and a backward pass (BP,
Algorithm \ref{alg:bsnew}) that
adds edges $e \in {\cal T} \times {\cal S}$ to ${\cal
G}$.
FP and BP  are essentially the
same. To abstract from the direction, we will use the terms
query language and retrieval language. In FP (resp.\ BP),
the query language is the source (resp.\ target) language and the retrieval
language is the target (resp.\ source) language.
\begin{itemize}
\setlength\itemsep{0.0em}
\item Let $q$ be the query string from the query language.
\item
The set $R$ holds retrieval language strings that are highly
associated with $q$. $R$ is initially empty.
$R$ is $T$ (a set of target strings) or $S$ (a set of English source strings) in the algorithms.
\item In each iteration,
we find the retrieval language string $r$ with the
highest association to those verses containing the query
$q$ that are not yet covered by $R$.
\item We terminate when coverage by $R$ (of verses
containing $q$) exceeds the threshold
$\alpha$. 
\item We return all edges that go from a query
language node that contains $q$ to a retrieval language node
that contains a string from $R$.
\end{itemize}
The formal description in \figref{fpbp} 
is slightly more complex because the query
$q$ is not a single string but a set. But this extension is straightforward.
We now explain the formal description in \figref{fpbp}.

We invoke FP and BP for all pairs (focal concept $F$,
target language $l$) and merge the result with $\cal G$ for
each invocation. Writing PASS for FP or BP:
\begin{align*}
{\cal G} \gets 
{\cal G} \cup \mbox{PASS}(F,l)
\end{align*}
For the following description of the algorithms, we write
$s \in v$ for ``string $s$ (in language $l$) is a substring
of (the language $l$ version of) verse $v$''.
For brevity, we describe FP (Algorithm 1) [and describe BP
(Algorithm 2) in square brackets].
Line 1: $T$
[$S$] collects target [source] strings.
Line 2: $M$
is the maximum number of iterations; we set $M =5$.
Line 3: $V$ is the set of verses that contain a string
in $F$ [were linked by an edge from $F$ in FP], but
are not yet covered by $T$ [$S$].  Line 4: We save the result for
$i=1$ (or  $T=\emptyset$ [$S=\emptyset$]) in $V_1$,
the
base set of verses.
Line 5: If the coverage that
$T$ [$S$] has of $V_1$ exceeds a threshold $\alpha$, we
terminate; we set $\alpha=.9$.  Line 6: We  find the
target string $t$ [source string $s$] that is most
associated with $V$, ignoring target [source] string
candidates already covered. Line 7: $t$
[$s$] is added to $T$ [$S$].  Line 9: In FP, we return a
set of new edges that start at the focal concept $F$ and end
at a target node $(l,v,T)$ whose verse $v$ contains a string
$t$ from $T$.  Line 9: In BP,
we return a set of new edges that start at a target
node $(l,v,T)$ that was connected to $F$ in FP
and end at an $S'$ that contains a highly associated source
string $s$ (i.e., $s \in S$) in $v$.

\section{Evaluation}


\subsection{Single concept across all languages}
We first evaluate how well our method performs at
identifying associated concepts across the highly diverse set of
languages we cover. Since there is no appropriate
broad-coverage high-quality resource,
this requires an expensive manual
analysis by a linguist.  We can therefore only perform it for
one concept in this paper. We choose the focal concept
\concept{bird}, defined as
\{\texttt{\$bird}, \texttt{\$fowl}, \texttt{\$flying\$creature}, \texttt{\$winged\$creature}\}.
For each language $l$, we analyze the
hits we get for
\concept{bird} in $l$, primarily by looking at its
BP hits in English, i.e., the English strings that are
proposed in BP by running \methodname
on \concept{bird}.
Defining
$R$ as the set of verses in which BP hits occur
and $B$ as the set of verses in which `bird' occurs,
we use
four evaluation categories.
(1) \textbf{one-to-one}. $ R \approx  B$. 
In detail:
$| R -  B| < .1 | B|$ and $ R -  B$
does not contain plausible additional  hits.
(2) \textbf{polysemy}.
$ R \supset  B$ and
$ R -  B$ consists of verses with
concepts closely related to \concept{bird}, e.g., 
\concept{dove}, \concept{fly}.
(3) \textbf{ambiguity}.
$ R -  B$  contains verses in which neither 'bird'
nor closely related meanings occur.
However, there is a
second ``non-bird'' meaning of the BP hits;
e.g., for Adioukrou the FP hit is
 ``{\textopeno}r'' and the BP hits
correspond to two clusters, a \meanings{bird} cluster and a
\meanings{hitting} cluster.
(4) \textbf{failure}.
$ R -  B$ or
$ B -  R$ is large and this cannot be attributed to
polysemy or (simple) ambiguity.
See
\secref{bird_first_eva} for details.
\tabref{bird_results} (top)  shows
that \methodname found the translation of \concept{bird} in almost
all languages where we count the \textbf{ambiguity} case
(e.g., Adioukrou ``{\textopeno}r'' meaning both \meanings{bird}
and \meanings{hitting}) as a success. The search failed for
4 languages ($4=1335-1331$) for which we have no verse that
contains \concept{bird} in English and 11 languages for many of which the
number of verses was small. Thus, \methodname requires a
large enough parallel corpus for good performance.

\begin{table}
\centering
\footnotesize
\setlength\tabcolsep{3pt}
\begin{tabular}{cccc|c}
        \toprule
        \textbf{1-1} & \textbf{polysemy} & \textbf{ambiguity} & \textbf{failure}& \textbf{total}\\
        \midrule
        687 & 579 & 54 & 11 & 1331\\
        \midrule
        \addlinespace
        \textbf{match} & \textbf{overlap} & \textbf{no overlap} & \textbf{no translation}& \textbf{total}\\
        \midrule
        488 & 192 & 457 & 194 & 1331\\
        \bottomrule
    \end{tabular}
\caption{\label{tab:bird_results_panlex}\label{tab:bird_results}
Evaluation of \methodname for \concept{bird}.
Top: Linguistic analysis. Bottom: PanLex results.
}
\end{table}

We also evaluate on PanLex
\citep{kamholz-etal-2014-panlex}, \url{http://panlex.org}.
Defining $P$ as the translations from PanLex and $ T$ as
the FP hits for \concept{bird}, we use
the following four categories.
(1) PanLex gives \textbf{no translation}. $P =\emptyset$.
(2) \textbf{no overlap}. ${ P} \cap { T}=\emptyset$.
(3) \textbf{overlap}.  $0<|P  \cap T| < |T|$.
(4) \textbf{match}. $|P  \cap T| = |T|$.
See \secref{bird_second_eva} for details.
\tabref{bird_results} (bottom) shows that for PanLex
languages, \methodname performs well on
$\approx 60\%$: $(488+192)/(488+192+457)$. In a qualitative
analysis, we found four  reasons for the 457 \textbf{no overlap} cases.
(i) A language has a very small corpus in PBC.
(Sparseness was  also the reason for failure
in \tabref{bird_results}, top). (ii) \methodname did find
correct translations of \concept{bird},
 but they are missing from PanLex.
 (iii)
There is a dialect/variety mismatch Bible vs PanLex (no
occurrence of
the PanLex translation in our corpus).
(iv) PanLex incorrectly translates through an intermediate language.
For example, since PanLex has no direct
translation of English \concept{bird} to 
Chorote Iyowujwa, it goes through Gimi `nimi' (which means
both \meanings{bird} and \meanings{louse}) and
returns Chorote Iyowujwa `inxla7a'. But `inxla7a'
only means \meanings{louse}. Another example is that PanLex translates \concept{bird}
as `San'
instead of the correct \citep{sampu05ngochang} `nghoq' for
Achang.
Thus, PanLex translations through the intermediate mechanism
are unreliable while our FP hit can find the correct
translation.


Taking the two evaluations together (manual analysis of BP
hits and comparison of FP hits to PanLex translations), we interpret the
results as indicating that \methodname reliably finds the
correct translation of the focal concept, but can fail
in case of data sparseness.

\begin{table}
\centering
\footnotesize
\begin{tabular}{lrrrr}
\toprule
model & \mc{partial} & \mc{strict} & \mc{relaxed} & \mc{FP} \\ 
\midrule
\methodname & 87.21 & 84.88 & 89.69 & 1.03\\
Eflomal 0 & 89.52 & 87.80 & 91.23 & 10.42\\
Eflomal 1 & 86.98 & 84.88 & 89.18 & 4.50\\
Eflomal 0.1 & 78.68 & 76.12 & 81.44 & 1.07\\
\bottomrule
\end{tabular}
\caption{\label{tab:swadesh_results}
Recall of proposed Swadesh32 translations $T$ on NoRaRe translations $N$, averaged over concept-language pairs. The score for a concept-language pair
is ${| T \cap  N|}/{| N|}$ (partial), 1 iff $| T \cap  N| = | N|$ (strict)
and 1 iff $|{ T} \cap { N}| \geq 1 $ (relaxed). FP: average false positives.
}
\end{table}

\subsection{Swadesh concepts}
We next evaluate on Swadesh32 (\secref{data}).
\tabref{bird_results} indicates that 
PanLex quality is low for many languages.
We therefore use
NoRaRe \citep{tjuka2022linking},
\url{http://norare.clld.org}.
We
use all 582 concept-language pairs for which NoRaRe gives a translation. 
For a concept-language pair,
let $ T$ be the proposed translations 
(from \methodname or Eflomal) and  $ N$  gold (from
NoRaRe).
Then we compute recall as
${| T \cap  N|}/{| N|}$. We
match two ngrams
if one is a substring of the other; e.g., 
\normalstring{oiseau} is correct for
\normalstring{oiseaux}.
For Eflomal \citep{ostling2016efficient}, we set $T$ to the
set of
target language words aligned with one of the
focal concept words (e.g.,
\{\texttt{\$belly\$}, \texttt{\$bellies\$}\}).
Eflomal 0, 1, 0.1 denotes
that we
only keep translations
whose frequency is $>0$, $>1$ 
and
$>
.1 \Pi(l, F)$, respectively.

\tabref{swadesh_results}
shows that \methodname's Swadesh32 translations have high
recall (roughly 85\% and higher, depending on the measure),
with few false positives (1.03).  For Eflomal, however, as
we restrict matches to high-precision matches (i.e., going from 0
to 1 and to .1), both recall and false positives (FP) drop.
Our interpretation is
that the alignments obtained by Eflomal
are noisy: Eflomal misaligns the focal concept with many
irrelevant words.  In contrast to \methodname, Eflomal
offers no good tradeoff.  This validates that we
use \methodname instead of standard aligners like Eflomal.
Most importantly, the evaluation on NoRaRe shows
that \methodname has high recall and produces few false
positives, which are prerequisites for further reliable
exploration/analysis.  See
\secref{swadesh_evaluation} for details of the evaluation (including an additional experiment in terms of coverage compared with Eflomal).


\begin{figure}
  \centering  \includegraphics[width=0.48\textwidth]{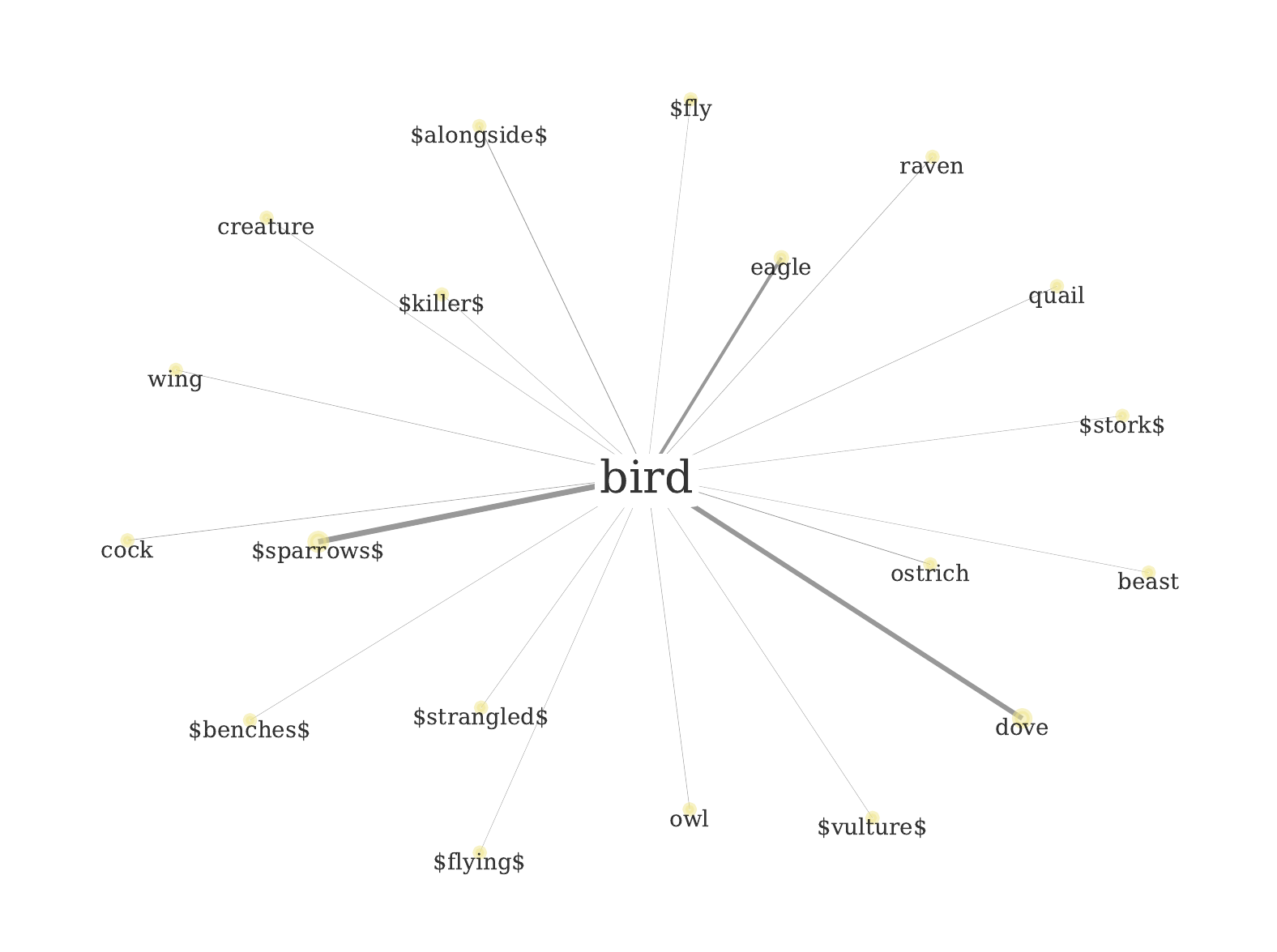}
  \caption{The crosslingual semantic field of `bird'}\label{fig:example}
\end{figure}

\subsection{Concept stability}
We define the crosslingual semantic field $\cal F$ of focal
concept $F \in {\cal  S}$ as
the second neighborhood of $F$,
the set of nodes at a distance 2 from $F$:
\begin{align*}
    {\cal F}(F) = \{ S  \in {\cal S}| \exists c: (F,c) \in {\cal G} \wedge
(c,S) \in {\cal G} \}
\end{align*}
\figref{example} shows the crosslingual semantic field of
`bird'. 
The strength of the line connecting \concept{bird} and
$S$ (which contains an English string) indicates the number of
languages through which \concept{bird} can reach $S$.
\normalstring{eagle}, \normalstring{dove} and \normalstring{sparrows} have thick lines,
indicating that there are many languages for
which \methodname
connects \concept{bird} to a target string whose meaning includes a
bird species. 
The size of a node indicates the number of paths from \concept{bird} to the string the node represents.
For example, the size of the \concept{bird} node (i.e., $F$) 
indicates the number
of recurrent paths,
i.e., $|\{c|(F,c) \in {\cal G} \wedge (c,F) \in {\cal G}\}|$.
The
visualization in \figref{example} suggests that \concept{bird} is \emph{stable}
crosslingually:
if we go roundtrip from English to a target
language $l$ and back,
in most cases what we get is \concept{bird}.
This is often not true (as we will see shortly) for a more
abstract concept like \concept{mercy}.
The proportion of recurrent paths is small: many paths starting from \concept{mercy} go to other nodes, such as \normalstring{pity} and \normalstring{poor},
indicating that it
is unstable. 
See \secref{crosslingualsemanticfields} for visualizations of all 83 concepts.

We define the \emph{stability} $\sigma(F)$ of a focal concept $F \in \cal S$ as:
\begin{equation*}
\sigma(F) = \frac{|\{ c | (F,c) \in {\cal G} \wedge (c,F) \in {\cal G}\} |}
{|\{ c  |  (F,c) \in {\cal G}\}|}
\end{equation*}
Thus, for a stable concept $F$ (one whose stability is close to 1.0), most paths starting from $F$ are part of a ``recurrent'' path that eventually returns to $F$. In contrast, an unstable concept $F$ like \concept{mercy} has relatively fewer such recurrent paths and a large proportion of its paths go to other concepts.

\begin{table}
\small
\centering
\begin{tabular}{cccc}
\toprule
 Accuracy & Precision & Recall & $F_1$  \\
\midrule
0.71 & 0.65 & 0.88 & 0.75 \\
\bottomrule
\end{tabular}
\caption{\tablabel{concreteness_stability_binary}
Performance of predicting a concept's stability from its concreteness
}
\end{table}

We hypothesize that one cause of stability is concreteness:
concrete concepts are more stable across languages than
abstract ones because they are directly grounded in a
perceptual reality that is shared across languages.
To test this hypothesis, we define a concept to be concrete
(resp.\ abstract) if its concreteness score $\gamma$ according to 
\citep{brysbaert2014concreteness} is $\gamma \geq 3.5$
(resp.\ $\gamma \leq 2.5$). 69 of our 83
concepts are either abstract or concrete, according to this definition
(see Tables \ref{tab:concept_stability1} and \ref{tab:concept_stability2} in the Appendix for concreteness and stability measures of all 83 concepts).
We define a concept $F$ to be stable iff $\sigma(F)\geq 0.6$.
\tabref{concreteness_stability_binary} shows that when we
predict stability based on concreteness (i.e., a concept is
predicted to be concrete iff it is stable), accuracy is
high: $F_1 = .75$.
This is evidence that our hypothesis is correct:
concreteness is an important contributor to stability.
See \secref{analysisconceptstability} for further analysis of the stability of concepts.

\subsection{Language similarity}
\seclabel{languagesimilarity}

\def\csimangle{75}
\begin{table}
\footnotesize
\centering
\begin{tabular}{rc@{\hspace{0cm}}|cccccc|c}
\rotatebox{\csimangle}{$k$}
& \rotatebox{\csimangle}{concepts}
& \rotatebox{\csimangle}{\textcolor{atla_color}{ATLA}}
& \rotatebox{\csimangle}{\textcolor{aust_color}{AUST}}
& \rotatebox{\csimangle}{\textcolor{indo_color}{INDO}}
& \rotatebox{\csimangle}{\textcolor{guin_color}{GUIN}}
& \rotatebox{\csimangle}{\textcolor{otom_color}{OTOM}}
& \rotatebox{\csimangle}{\textcolor{sino_color}{SINO}}
& \rotatebox{\csimangle}{all}\\
\toprule
\multirow{3}{*}{2} & 32 
& .21  & .2  & .53  & .09  & .14  & .00 &.13\\
                     & 51
                     & .24  & .19  & .26  & .08  & .04  & .03 &.11\\
                     & 83
                     & .29  & .31  & .49  & .11  & .14  & .04 &.17\\\midrule
\multirow{3}{*}{4} & 32 
& .54  & .41  & .80  & .24  & .39  & .15 &.29\\
                     & 51
                     & .52  & .45  & .48  & .18  & .12  & .09 &.24\\
                     & 83
                     & .63  & .51  & .77  & .31  & .28  & .09 &.32\\ \midrule
\multirow{3}{*}{6} & 32 
& .63  & .49  & .85  & .30  & .43  & \underline{.16} &.33\\
                     & 51
                     & .64  & .57  & .57  & .20  & .13  & .13 &.30\\ 
                     & 83
                     & .74  & \underline{.60}  & .83  & .40  & .37  & .12 &.37\\ \midrule
\multirow{3}{*}{8} & 32
& .68  & .53  & \textbf{.87}  & .34  & \underline{.51}  & \textbf{.18}&.36 \\
                     & 51
                     & .71  & .59  & .60  & .22  & .14  & .15 &.32\\
                     & 83 & \underline{.78}  & \underline{.60} & \underline{.86}  & \textbf{.42} & .36 & \textbf{.18} &\textbf{.39}\\
                     \midrule
\multirow{3}{*}{10} & 32
& .73  & .56  & .84  & .34  & \textbf{.54}  & \textbf{.18} &.37\\
                     & 51
                     & .74  & \textbf{.61}  & .61  & .21  & .09  & .12 &.32\\
                     & 83
                     & \textbf{.80}
                     & \textbf{.61}  & .83
                     & \underline{.41}  & .28
                     & \underline{.16} & \underline{.38}
\end{tabular}
\caption{\label{tab:language_family_classification1}
Accuracy of prediction of typological family based on nearest
neighors in \methodname-based representation space.
Representations for Swadesh32 (32), Bible51 (51)
and All83 (83) concepts. $k$: number of nearest
neighbors. Family abbreviations: see text. 
\textbf{Bold} (\underline{underlined}): best (second-best)
result per column.
}
\end{table}

We now propose and evaluate a new measure of 
similarity between languages,
\emph{conceptual similarity},
based on conceptualization. Since
we have aligned concepts across languages, we can compute
measures of how similar the conceptualization of two
languages is. For example, in contrast to Western European
languages, Chinese, Korean, and Japanese have one concept
that means both \meanings{mouth} and \meanings{entrance}. Our measure aggregates
such patterns over many concepts and predicts
higher similarity between the three East Asian languages and
lower similarity to Western European languages.

To compute conceptual similarity, we represent a language
$l$ as the concatenation of 83 vectors $\vec{v}(l,F_j)$, each capturing how
it represents one of our 83 concepts:
\begin{align*}
    \vec{v}(l) = [\vec{v}(l,F_1);\vec{v}(l,F_2);\ldots;\vec{v}(l,F_{83})]
\end{align*}
where $[;]$ is vector concatenation. 
We define $\vec{v}'(l,F_j)$ as a 100-dimensional vector and
set
\begin{align*}
    \vec{v}'(l,F_j)_i=
|\{ c | (F_j,c) \in {\cal G} \wedge (c,\{e_i\}) \in {\cal
G}|
\end{align*}
i.e., the number of paths from $F_j$ to the English ngram
$e_i$; here we only consider nodes $c=(l',v,T)$ for which
$l'=l$, i.e., only nodes that 
belong to language $l$. 
For example, \concept{mouth} connects with Chinese nodes
 containing \normalstring{\begin{CJK*}{UTF8}{bsmi}口\end{CJK*}} in FP. BP
 connects these nodes not only
 to \concept{mouth}, but also to \normalstring{entrance}.
Our convention is that the first dimension 
$\vec{v}'(l,F_j)_1$ always
represents the value of the focal concept $F_j$.
To define the other dimensions,
we sort all
associated English ngrams $e_k$ 
according to 
the number of languages in which
they are associated with $F_j$ and select the top 99\footnote{For some focal concepts that are less divergent, e.g., \concept{bird}, we obtain fewer than 99 dimensions};
these are then the dimensions 2-100 of $\vec{v}'(l,F_j)$.
We compute the final
vector $\vec{v}(l,F_j)$
by normalizing
$\vec{v}'(l,F_j)$ by
$\sum_k \vec{v}'(l,F_j)_k$.

$\vec{v}(l,F_j)$
captures which concepts related to $F_j$ are
clustered in $l$ and thereby indicates $l$'s similarity
to other languages. For example, for
the focal concept \concept{mouth},
the $\vec{v}(l,F_j)$ for Chinese, Japanese and Korean are
more similar, but they are less similar
to $\vec{v}(l,F_j)$
for Western European
languages. 

We can now define the \emph{conceptual similarity}
between two languages $l_1$ and $l_2$ as the cosine
similarity between their vectors:
\begin{align*}
    \mbox{c-sim}(l_1,l_2) = \cos(\vec{v}(l_1),\vec{v}(l_2))
\end{align*}

We evaluate on
Glottolog
4.7 \cite{harald_hammarstrom_2022_7398962}.
We select the six language families that
have more than 50 members in the PBC:
\textcolor{atla_color}{Atlantic-Congo (ATLA)},
\textcolor{aust_color}{Austronesian(AUST)},
\textcolor{indo_color}{Indo-European (INDO)},
\textcolor{guin_color}{Nuclear Trans New Guinea (GUIN)},
\textcolor{otom_color}{Otomanguean (OTOM)} and
\textcolor{sino_color}{Sino-Tibetan (SINO)}. 
We then evaluate conceptual similarity on a
binary classification task:
\emph{Is the
majority of  language $l$'s $k$ nearest neighbors
in the same family as $l$?} In addition to representations
based on all 83 focal concepts (referred to as \textbf{All83}), we also analogously create
representations based just on Swadesh32 and Bible51.

\tabref{language_family_classification1} shows
that for two ``dense'' families (i.e., most members have
close relatives), our results are good (up to .8 for
\textcolor{atla_color}{ATLA}, .87 for \textcolor{indo_color}{INDO}). For \textcolor{aust_color}{AUST}, \textcolor{guin_color}{GUIN} and \textcolor{otom_color}{OTOM}, about half of
the predictions are correct for the best $k$. \textcolor{sino_color}{SINO}
performance is bad, indicating that \textcolor{sino_color}{SINO} languages are
conceptually more distant from each
other.
The difference
between Swadesh32 and Bible51 performance is large in some
cases, especially for
\textcolor{indo_color}{INDO} and \textcolor{otom_color}{OTOM}.
We hypothesize
that the conceptualization for more
abstract concepts in Bible51 
is more variable than for more
concrete concepts in Swadesh32.

The main result of this evaluation is
that the language representations (i.e., the $\vec{v}(l)$)
derived by \methodname 
are a good basis for assessing the similarity of languages.
\secref{clustering} discusses in detail the
complementarity of conceptual similarity to other similarity
measures.

\begin{figure}
  \centering
  \includegraphics[width=0.48\textwidth]{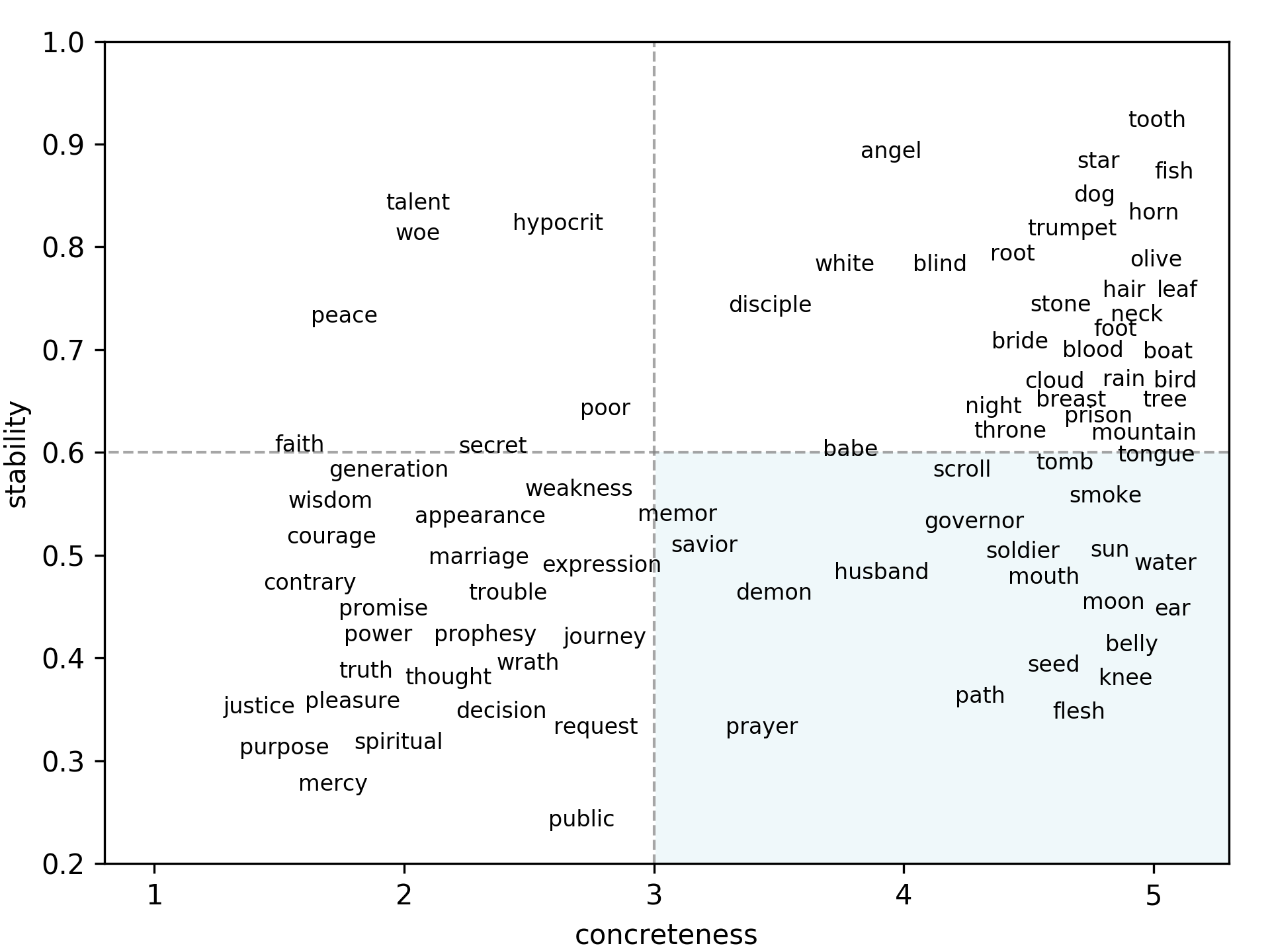}
  \caption{\textit{stability} measure vs. \textit{concreteness} scores.}\figlabel{stability_concreteness}
\end{figure}

\section{Analysis}
\subsection{Concept stability}
\seclabel{analysisconceptstability}
Figure \ref{fig:stability_concreteness} and
Table \ref{tab:concreteness_stability_binary} support our
hypothesis that concreteness contributes to stability,
as most concrete concepts tend to be stable and most abstract concepts tend to be less stable.
However,
recall is higher than precision in
Table \ref{tab:concreteness_stability_binary}, indicating
that stable concepts are mostly concrete but not vice
versa, as plenty of concepts are located in the lower right in Figure \ref{fig:stability_concreteness}. In our analysis, we find that some concrete concepts
are unstable because meaning is extended by
semantic processes
like
metonymy and metaphor.
We now give examples of our analysis
of this phenomenon in \tabref{unstability}.
See \secref{crosslingualsemanticfields} for
the visual semantic fields 
of all 83 concepts.

\tabref{unstability} 
shows (i) concrete concepts whose instability is plausibly
caused by
metaphor, metonymy, and other semantic extensions and (ii)
crosslingually associated English concepts that indicate instability.
For example, metonymy, where a concept is represented by the name of something associated with it, results in instability frequently. This is in line with crosslinguistic studies that show that metonymy seems to be a universal language phenomenon \citep{khishigsuren2022metonymy}.

\begin{table}
\centering
\small
\setlength\tabcolsep{3pt}
\begin{tabular}{lll}
        \toprule
        types & concept & associated \\
        \midrule
        &`ear' & \textit{hear}, \textit{listen} \\
        & `mouth' & \textit{word}, \textit{speak} \\ 
        \multirow{1}{*}{\textbf{metonymy}} & `sun' & \textit{day}, \textit{east}\\
        & `moon' & \textit{month}\\
        & `belly' & \textit{womb}, \textit{birth}\\
        & `soldier' & \textit{army}, \textit{military}\\
        \midrule
        \multirow{2}{*}{\textbf{metaphor}} & `seed' & \textit{son}, \textit{offspring}\\
        & `mouth' & \textit{entrance}\\
        \midrule
        \multirow{2}{*}{\textbf{other}} & `water' & \textit{waters}, \textit{sea}, \textit{drink} \\
        & `knee' & \textit{obeisance}, \textit{worship}\\
        \bottomrule
    \end{tabular}
\caption{\label{tab:unstability}
Selected examples of focal concepts that are unstable due to different types of meaning extensions: \textbf{metonymy}, \textbf{metaphor}, and \textbf{other}.
}
\end{table}

Processes of semantic extension are central to cognitive
linguistics for explaining historical changes, language use
and framing in discourse
\cite{evans2006cognitive,lakoff2008metaphors}.
Since these processes are pervasive 
in communication \citep{sopory2002persuasive}, in the lexicon \citep{khishigsuren2022metonymy} and in historical
language change \citep{xu2017evolution},
they are also a plausible cause for crosslingual instability of concepts.

\begin{figure*}
\begin{tabular}{c}
\hspace{-0.4cm}
\includegraphics[width=0.45\textwidth]{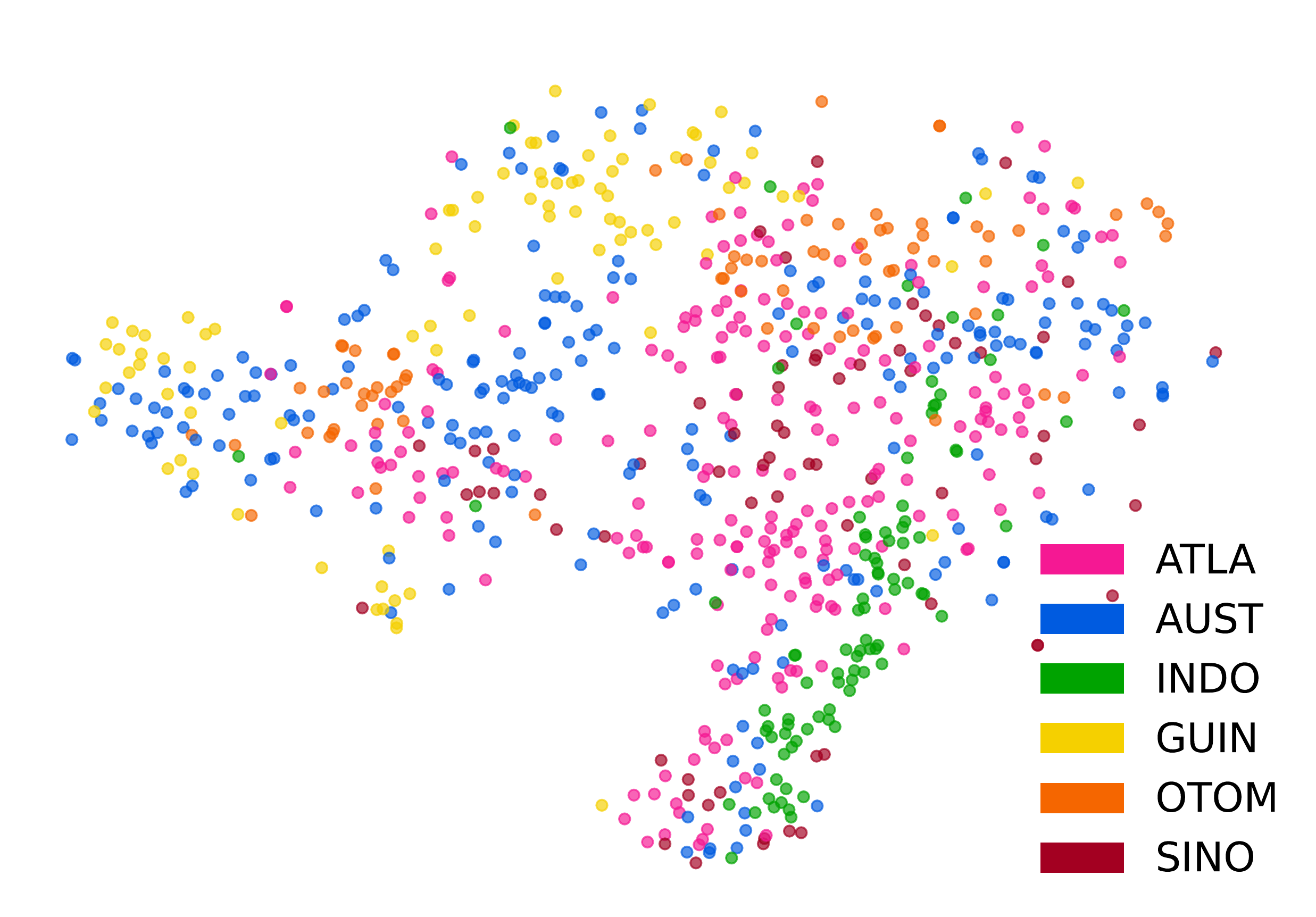}
\hspace{0.8cm}
\includegraphics[width=0.45\textwidth]{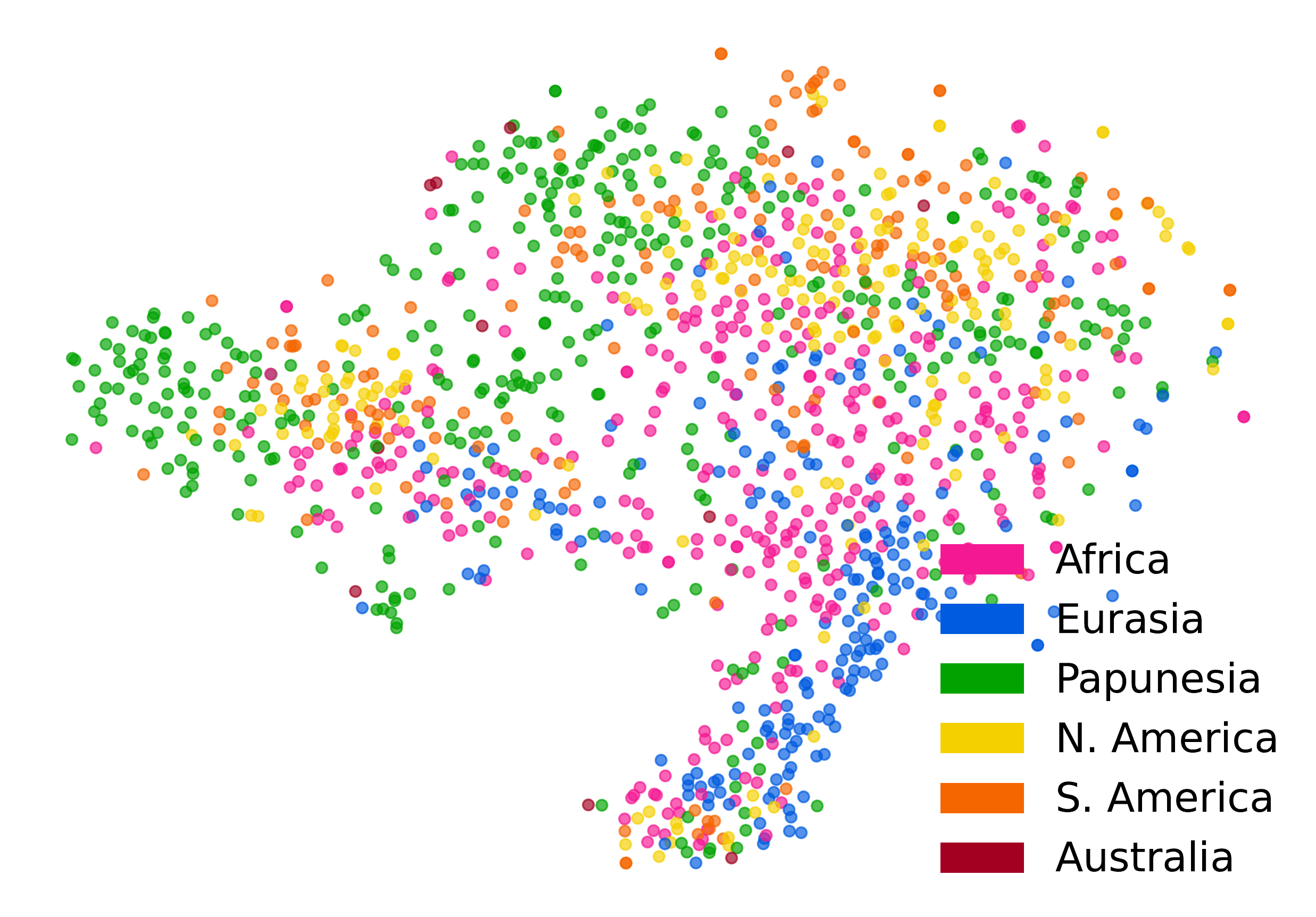}
\end{tabular}
\caption{t-SNE  of  languages represented
as \methodname-based vectors of the Swadesh32 concepts.
Colors indicate family (left) and area (right).}
\label{fig:clustering}
\end{figure*}

\subsection{Language similarity}
\seclabel{clustering}
\figref{clustering}
shows a visualization of 
conceptual similarity using
t-SNE \citep{van2008visualizing}.
Each point in the figure
represents a language. Color indicates family
on the left
(for the six families from
\secref{languagesimilarity})
and area on
the right (for all languages, area extracted
from Glottolog 4.7).

The right figure (area) shows that conceptual similarity is
correlated with geographic proximity: except for
\textcolor{aus_color}{Australia},
all regions are concentrated in a few subclusters, with
\textcolor{eur_color}{Eurasia} being most clustered and \textcolor{afr_color}{Africa} least.
For example,
there are two
clusters (top left, top center) that cover most \textcolor{sam_color}{South
American} languages. This is
consistent with findings  that close languages 
often influence each other \citep{sapir1912language}.
Of course, the
main reason is  the correlation
between
conceptual and typological relatedness (see below) and
between typological and geographic proximities.
But there are also indications of geography-induced
conceptual similarity, e.g., 
most \textcolor{eur_color}{Eurasian}
languages have close \textcolor{afr_color}{African} neighbors.
We present examples at the end of \secref{clustering}.

That geographic proximity is not sufficient for conceptual
similarity is demonstrated by
\textcolor{pap_color}{Papunesia}: it is distributed over most of the 
figure, indicating that the underlying conceptualizations
are quite different. This is consistent with the well-known fact that
\textcolor{pap_color}{Papunesia} exhibits enormous linguistic diversity in a very
small geographic area \citep{foley1986papuan}.
Of course, there is still
subcluster structure:
most \textcolor{pap_color}{Papunesian} languages have near
\textcolor{pap_color}{Papunesian}
neighbors in
conceptual space.

Turning to the left figure (family),
we see that there is agreement of conceptual similarity
with typological similarity.
Some 
families form a relatively clear cluster, in particular,
\textcolor{indo_color}{Indo-European}, suggesting
that \textcolor{indo_color}{Indo-European} languages are similar in conceptualization.
This explains \textcolor{indo_color}{Indo-European}'s high accuracy in \tabref{language_family_classification1}.

But there is also complementarity between conceptual
similarity and typological similarity.
\textcolor{sino_color}{Sino-Tibetan}
is spread out across
the entire figure, explaining its low accuracy
in \tabref{language_family_classification1}.
For Chinese and Tibetan,
we find their conceptualizations to be quite different, in
particular, for
body parts
(e.g., \texttt{mouth} and \texttt{neck}). See \secref{furtheranalysislangsim} for examples.

Conversely, we now present three examples of typologically
distant languages that are still conceptually close.
The first example is \textbf{Tagalog}.
While \textcolor{indo_color}{Indo-European} languages
mostly occur in a relatively tight region of the figure,
that region also contains many non-Indo-European
languages. We hypothesize that \textcolor{indo_color}{Indo-European} languages have
influenced the conceptualization of other languages
worldwide due to their widespread use, partly as part of
colonization. One example is that the Tagalog
words
\normalstring{dila} and \normalstring{wika}
mean both \meanings{tongue}
and \meanings{language}, a conceptualization similar to Spanish
(\normalstring{lengua}), a language Tagalog
was in close contact with for centuries. Standard Malay is
typologically related to Tagalog, but its word
for \meanings{tongue} \normalstring{lidah}, does not include the
meaning \meanings{language}. This may contribute to Tagalog being
conceptually more similar to Spanish on our measure than other \textcolor{aust_color}{Austronesian}
languages. 
\textbf{Plateau Malagasy},
an \textcolor{aust_color}{Austronesian} language spoken in Madagascar, is
conceptually similar to both
far-away \textcolor{aust_color}{Austronesian} languages like Hawaiian (reflecting its typology) as well as
to geographically close, but typologically dissimilar
\textcolor{atla_color}{Atlantic-Congo} languages like Mwani  and Koti.
\textbf{Masana} is an Afro-Asiatic language spoken
in Nigeria. It is conceptually close to the \textcolor{atla_color}{Atlantic-Congo}
languages 
Yoruba, Igbo 
and Twi, also spoken in and around Nigeria.
Geographic proximity seems to boost conceptual similarity in
these three cases.
We leave further investigation of
the hypothesis that \methodname-based representations reveal
historical interlanguage influences to future work.

\section{Conclusion \& Future Work}
We propose \methodname, a method that automatically aligns
source-language concepts and target-language strings by
creating a directed bipartite graph. We investigate the
structure of such alignments for 83 focal concepts. Our
extensive manual evaluation demonstrates good performance
of \methodname.  We introduce the notion of crosslingual
stability of a concept and show, using \methodname, that
concrete concepts are more stable across languages than
abstract concepts.  We also define conceptual similarity, a
new measure of language similarity based on \methodname
representations. In our experiments, conceptual similarity
gives results that partially agree with established measures
like typological and areal similarity, but are complementary
in that they isolate a single clearly defined dimension of similarity:
the degree to which the conceptualization of two languages
is similar. 

In the future, we would like to improve the
efficiency of \methodname and,
extending our work on a sample of 83 in this paper,
apply it to all concepts that occur in PBC.

\section*{Limitations}
The \methodname we propose consists of two core steps, i.e., forward pass and backward pass. The forward pass identifies the most associated target-language strings for a focal concept. However, due to possible data sparsity of PBC in some low-resource languages and some cases of verse-level misalignment, $\chi^2$ scores of the real translations can be indistinguishable compared with some other rare words that also occur in the same verses. Under such rare cases, \methodname will not work well enough. In addition, the genre of PBC is limited to religion and therefore the diversity of the concepts across languages is largely influenced. Nevertheless, PBC, as far as we know, provides texts in the largest number of low-resource languages. PBC is thus a good fit for our goal.

In this work, we select 83 concepts, including the Swadesh32 and Bible51, representing a wide range of interesting crosslingual concepts. The runtime for computing the results for one concept in all languages is around 10 hours on average. The relatively long runtime, however, can prevent us from exploring more interesting concepts.

We find that the concreteness of a focal concept can be a contributor to the stability measure. As we use English as the source language for representing the focal concepts, we naturally resort to concreteness scores from English language ratings only. In addition, the analysis is carried out from an English perspective. Nevertheless, as we want to compare different languages, we have to use a unified source language. Theoretically, we can use any language as the source language and represent the concepts in that language. We therefore plan to use other languages, e.g., Chinese, or some low-resource languages, as the source language in future research.

\section*{Ethics Statement \& Risks}
In this work, we investigate the differences in conceptualization across 1,335 languages by aligning concepts in a parallel corpus. To this end, we propose \methodname, a method that creates a directed bipartite alignment graph between source language concepts and sets of target language strings. The corpus we used, i.e., PBC, contains translations of the Bible in different languages (one language can have multiple editions). As far as we know, the corpus does not include any information that can be used to attribute to specific individuals. Therefore, we do not foresee any risks or potential ethical problems.  

\section*{Acknowledgments}
We would like to acknowledge Verena Blaschke for her valuable suggestions. We would also like to thank the reviewers for
their positive and constructive feedback. This work was funded by the European Research Council (grant \#740516).

\bibliography{anthology,custom}
\bibliographystyle{acl_natbib}

\appendix

\section{Details of data}\seclabel{additional_details_data}
\subsection{Parallel Bible Corpus}\seclabel{details_pbc}

We work with the Parallel Bible Corpus
(PBC, \citep{mayer-cysouw-2014-creating}), which contains
1,775 editions of the Bible in 1,335 unique languages (we
regard dialects with their own ISO 639-3
codes\footnote{https://iso639-3.sil.org/} as different
languages). As far as we know, there is no explicit licence for PBC dataset. For each language, we only use one of its
available editions. We use the \textit{New World} edition
for each language, if available, and the edition with the largest number of verses otherwise. Different from previous
work \citep{asgari-schutze-2017-past,
dufter-etal-2018-embedding, weissweiler-etal-2022-camel}
which only used verses that are available in all
languages, we use all parallel verses between English and
any other target languages. This means the number of
parallel verses between English and other languages can be
different. In general, we have parallel verses with a number
greater than 30,000 (Hebrew + New Testament) for high-resource
languages (141 languages), e.g. French, German and Chinese, while around
7,900 (New Testament only) for most of the other languages (1038 languages). 

\subsection{Concept selection}\seclabel{details_concept_select}

We have 83 focal concepts in total which are listed in Tables \ref{tab:concept_stability1} and \ref{tab:concept_stability2}.
We classify the focal concepts into Swadesh32 and Bible51, for which we explain the selection of concepts in detail as follows.

The \textit{Swadesh 100} \citep{swadesh2017origin} list offers 100 English words that represent universal and basic concepts. The words include nouns, adjectives, and verbs. We limit our selection to nouns to facilitate the comparison between concepts.
Because we choose Bible editions as our resource, many concepts in the list can have very low frequencies of occurrences or even do not occur at all. On the contrary, some concepts in the list can occur many times but are less interesting to us, e.g., `I', `you', `we', `this' etc. We therefore only keep the concepts in the list which occur equal to or more than 5 times in the New Testament (only New Testament is available for most low-resource languages so the concept has to appear in the New Testament) and less than or equal to 500 times in Hebrew + New Testaments. There are 32 concepts that fulfill the criterion and we refer them to as \textbf{Swadesh32}, which are shown in Table \ref{tab:concept_stability1}.

For Bible concepts, we first obtain the distinct types of strings that have a length between 4 and 15 characters from all words in the English Bible. The ngrams can contain but cannot go beyond the word boundaries. For example, from a part of sentence \texttt{\$bird\$fly\$} (substituting whitespaces with \texttt{\$}), we can obtain ngrams such as \texttt{\$bird}, \texttt{\$bird\$} and \texttt{fly\$}, but \texttt{\$bird\$f} is not possible because it contains \texttt{\$} in the middle. After this, we randomly select 10 languages from the available languages of PBC plus Chinese and German (12 languages in total). For each of the selected languages, we compute the coverage of the identified most associated target-language string obtained by performing a forward pass (setting the max number of iterations $M$ to 1) (Algorithm \ref{alg:fs}) by regarding each English string as a focal concept. If the coverage of a string is larger than 0.5 for more than five languages, then we keep it otherwise we filter it out. This results in around 1000 strings. Then we filter out those that represent named entities. Finally, we manually check the list and select the strings that represent nouns and are not in the Swadesh list and are more or less specific to the Bible. This finally results in a set of 51 concepts (\textbf{Bible51}) which are shown in Table \ref{tab:concept_stability2}.

\section{Additional details of \methodname}\seclabel{additional_details}

\subsection{Focal concepts \& strings}
The bipartite graph $\cal G$ we construct contains source nodes set $\cal S$ and target nodes set $\cal T$. Each node $s \in \cal{S}$ is a concept and is represented by a set of strings. We restrict the length of the strings between 1 and 8 for any language except the source language English (we restrict the length larger than 2 and the strings cannot go beyond word boundaries) for efficiency. To differentiate the nodes in  ${\cal S}$, we refer to the set of our chosen 83 focal concepts as ${\cal S}_F \subset {\cal S}$. Each focal concept $F$ in ${\cal S}_F$ is a set which can contain multiple strings, e.g., `belly' concept: $\{\texttt{\$belly\$}, \texttt{\$bellies\$}\}$. In contrast, other concepts in ${\cal S}$ are sets which contain only a single string, e.g., $\{\texttt{\$sparrows\$}\}$. In the backward pass for a focal concept $F$, if a string $s$ being identified belongs to $F$, then we create an edge that ends at $F$ instead of ${s}$. 

\subsection{String candidates}
The \methodname consists of (1) a forward pass: for building edges $(F, c) \in {\cal S} \times \cal{T}$ and (2) a backward pass for building edges $(c, f) \in {\cal T} \times {\cal S}$. In the forward pass, for example, an edge $(F, c) \in {\cal S} \times {\cal T}$ is constructed if a target node from the target language $l$: $c = <l, v, T>$'s verse $v$  contains a string $t$ that is highly associated with $F$. As the search space of target-language strings is extremely large for each $l$, we therefore restrict the search space to the set of strings which occur in the verses whose corresponding English verses contain $F$. Formally, let $\Pi(\text{eng}, F)$ be the verses where the focal concept $F$ occurs and ${\cal P}(\Sigma^*)(l, v)$ be the strings that fulfill our string selection conditions in verse $v$ for language $l$. We will then only consider the strings in $\mathbb{T} = 
\bigcup_v {\cal P}(\Sigma^*)(l, v) | v \in \Pi(\text{eng}, F)$
to be candidates in language $l$ that are possibly associated with $F$. Similarly, in the backward pass, we also restrict the search space to be the set of English strings in the verses whose corresponding target-language verses contained the identified target-language string set $T$ in the forward pass. Formally, let $\Pi(l, T)$ be the verses where the $T$ occurs and ${\cal P}(\Sigma^*)(\text{eng}, v)$ be the strings that fulfill our string selection conditions in verse $v$ for English.  We will then only consider English strings in $\mathbb{S} = 
\bigcup_v {\cal P}(\Sigma^*)(\text{eng}, v) | v \in \Pi(l, T)$.  Furthermore, we will only consider the strings $t \in \mathbb{T}$ which occur more than $\Pi(\text{eng}, F) / 10$ in the target-language verses of $\Pi(\text{eng}, F)$ and $s \in \mathbb{S}$ which occur more than 2 times in the English verses of $\Pi(l, T)$.

\subsection{Measuring association}
Given a set of strings $U$ in a language $l_2$, we want to find a string in language $l_1$ that is associated with $U$.
 To this end, we use $\chi^2$ score to measure the degree of the association. Specifically, we divide the verse set into two subsets: verses containing $U$ and verses not containing $U$ in the Bible of $l_2$, i.e., $\Pi(l_2, U)$ and $\neg \Pi(l_2, U)$. We then build a contingency table for each string candidate $t \in {\cal P}(\Sigma^*)$ and $t$ comes from language $l_1$, as shown in Table \ref{tab:cont}. After that, we compute the $\chi^2$ score for each string: the higher the score, the more associated the string in $l_1$ is with the set of strings $U$. We then choose the string that has the highest $\chi^2$ score as a hit in language $l_1$ for a set of strings $U$ in language $l_2$.

\begin{table}
\centering
\footnotesize
\setlength\tabcolsep{3pt}
\begin{tabular}{ccc}
\hline
 & in $\Pi(l_2, U)$ & in $\neg \Pi(l_2, U)$ \\
\hline
$I(t)$ & $n_{00}$ & $n_{01}$ \\
$\neg I(t)$ & $n_{10}$ & $n_{11}$ \\
\hline
\end{tabular}
\caption{\label{tab:cont}
The contingency table for a given string $t$ from language $l_1$. $I(t)$ (resp. $\neg I(t)$) denotes the string occurs (resp. does not occur) in the corresponding parallel verses for $l_1$. For example, $n_{00}$ denotes the number of verses in $\Pi(l_2, U)$ in which $t$ occurs.
}
\end{table}

\subsection{Adding edges}
For efficiency and stability reasons, in the actual implementation of the backward pass, \methodname does not add edges from a target node to all the strings in $S$ that fulfill the criterion, i.e., the set of the identified associated source language strings, as shown in Algorithm \ref{alg:bsnew} (line 9). Instead, we only add one edge only from each target node. This means that in each iteration of the backward pass, we will add new edges starting from the involved target nodes to a single $s$: $\{((l,v,T),s)|(F, (l,v,T)) \in {\cal G} \wedge s \in v\ \wedge v \in V\}$. In this way, each target node that was previously connected with the focal concept $F$ can only be connected to one source node only. By doing this, we find that some undesirable associations can be avoided.

\subsection{Hyperparameters}
We have two hyperparameters in \methodname, i.e., (1) the maximum number of iterations of searching associated strings for a focal concept in each language: $M$ and (2) the threshold $\alpha$ for the minimum coverage of the set of identified associated string $U$. We set $M=5$ and $\alpha = .9$ as default values for all involved computations. Based on preliminary experiments on `bird' concept, we found that the number of associated strings usually will not go beyond 5. Moreover, when $M$ is large, we might have an efficiency problem (more iterations for each language) when computing other focal concepts. Therefore, we set $M=5$ to reduce the runtime of \methodname while not sacrificing the accuracy too much. As for the coverage threshold $\alpha$, we conduct preliminary experiments with .85, .9 and .95 respectively on `bird' concept. We found when coverage is small (<.9), the search stops when there are still possibly unidentified associated strings in the rest verses. If the coverage is too large (>.9), we found that some less related strings can be identified at the later stage of iterations for some languages. We should note that PBC can have verse-level misalignment problems, which means for some parallel verses `bird' occurs in English but the target-language verse can be unrelated to `bird' at all. Moreover, as we remove the parallel verses that we have covered in each iteration, the verses uncovered become smaller and smaller in each iteration. $\chi^2$ scores computed on later iterations can be not significant and multiple strings can have the same highest $\chi^2$ scores if they only appear in the uncovered verses with the same number of occurrences. Therefore, to ensure that \methodname finds enough strings while guaranteeing the quality of the associations between them with the search string, we set the coverage threshold $\alpha = .9$.

\begin{figure}
  \centering
  \includegraphics[width=0.48\textwidth]{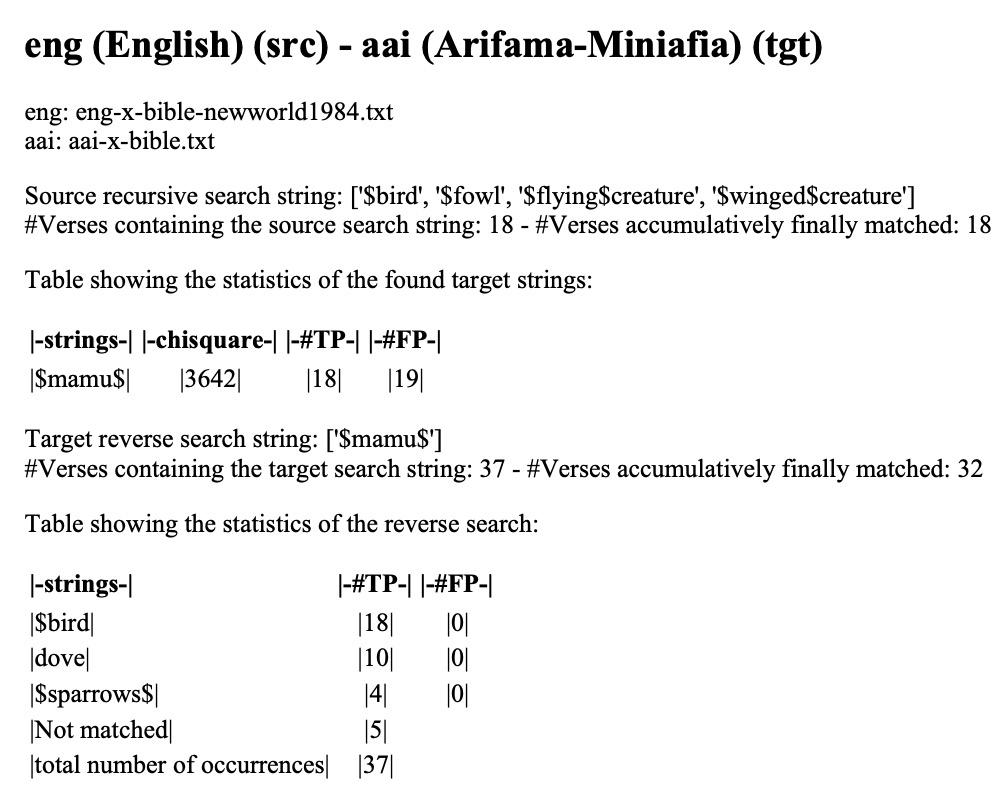}
  \caption{Part of the document for annotation for one target language, i.e., aai (Arifama-Miniafia) The true positive and false positive verses are not shown here.}\label{fig:annotation}
\end{figure}

\begin{figure*}
  \centering
  \includegraphics[width=1\textwidth]{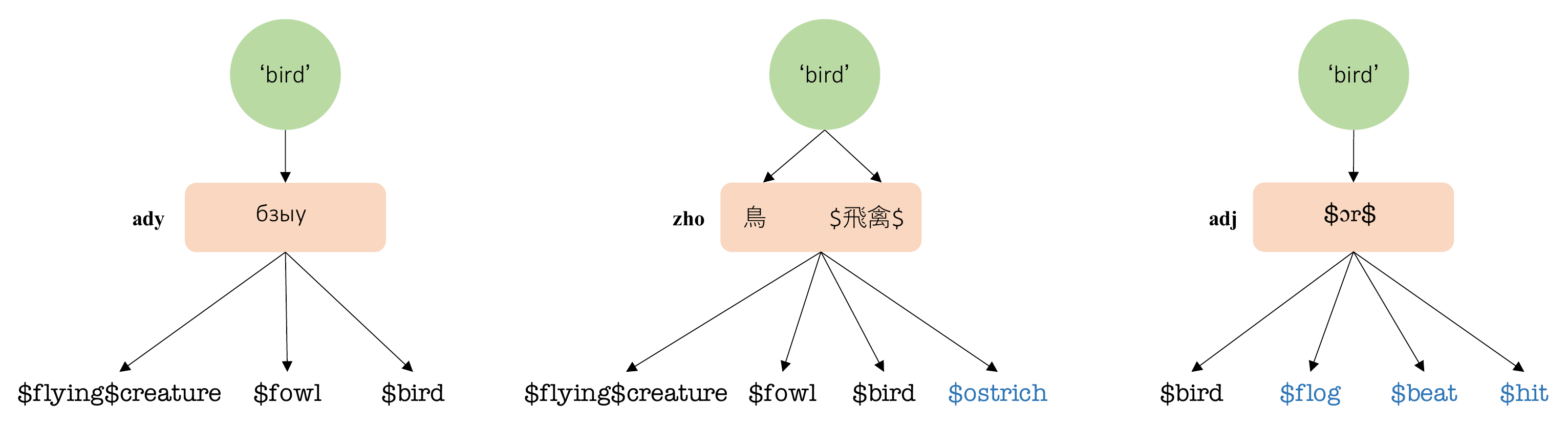}
  \caption{Examples of languages identified as \textbf{one-to-one}, \textbf{polysemy} and \textbf{ambiguity} respectively. The strings that do not belong to the set of strings representing the focal concept `bird' are marked with a different color. For Chinese, we have a BP hit of `ostrich', a hyponym of `bird'. For Adioukrou, we do not have BP hits that are closely related to `bird' but we have hits that are related to another cluster of meanings, i.e., \meanings{hit}.}\label{fig:categories}
\end{figure*}

\section{Details of Evaluation }\seclabel{details_evaluation}

\subsection{Single concept across all languages}\seclabel{bird_evaluation}

In the English version of the Bible, we find the `bird' concept is often expressed by the following words/phrases: ``bird(s)'', ``fowl(s)'', ``flying creature(s)'' and ``winged creature(s)''. Therefore, we use the following strings: \texttt{\$bird}, \texttt{\$fowl}, \texttt{\$flying\$creature} and \texttt{\$winged\$creature} to represent the `bird' concept. We agree that this set of strings might not be optimal for other large-resource parallel datasets (if there are any). However, for PBC dataset, this set of strings could empirically cover the `bird' concept.  We perform our \methodname which includes the forward pass and backward pass for this `bird' concept for all the languages in PBC. We perform three types of evaluations as follows based on the FP and BP hits:

\subsubsection{`bird' conceptualization in all languages}\seclabel{bird_first_eva}
We provide a pdf document in which the statistics of each string identified in both forward pass and backward pass are shown. In addition, for each target-language string, a) two randomly sampled \textbf{True Positive} parallel verses, i.e., target-language verses that contain the identified strings and the parallel English verses contain `bird'; b) two randomly sampled \textbf{False Positive} parallel verses, i.e., target-language verses that contain the strings and the parallel English verses that do not contain `bird', are shown. We also show three randomly sampled \textbf{False Negative} parallel verses, i.e., target-language verses that do not contain the strings and the parallel English verses that contain `bird' concept. An example of part of the document for one language is shown in Figure \ref{fig:annotation}. By checking the general patterns demonstrated in the document, 
we define four evaluation categories: 
\textbf{one-to-one},
\textbf{polysemy},
\textbf{ambiguity} and 
\textbf{failure}.
Noticeably, our category \textbf{polysemy} and \textbf{ambiguity} do not directly correspond to the definition in linguistics, but reflect general patterns of the conceptualization. The classification of these two categories is based on the pattern of strings identified in the backward pass. 
More specifically, we classify the conceptualization pattern of a language as \textbf{polysemy}, if it shows one of the following patterns: 
(a) \textit{hyponymy}, where we found strings such as \texttt{dove} and \texttt{sparrows}, which are hyponyms of `bird'.
(b) \textit{meronymy}, where we found strings like \texttt{\$wings}, which is a meronym of `bird'.
(c) \textit{other related words}, where we found strings such as \texttt{\$fly} and \texttt{\$chirp}, which are apparently related to `bird', but do not fit into any well-defined lexical semantic relation. The conceptualization pattern is classified as \textbf{ambiguity}, if the strings we found in the backward pass are not semantically related to `bird' at all (such as \texttt{\$new\$} and \texttt{\$kid\$}), but nevertheless are deemed as highly associated with `bird' by our algorithm. These cases are generally caused by having homonyms of `bird' in the target language. In case the linguist annotator cannot be sure of the classification, consultation has been made with other experts to resolve these issues and find the most common agreement.

\begin{table}
\centering
\footnotesize
\setlength\tabcolsep{3pt}
\begin{tabular}{llll}
\hline
 $l$ & ${P}$ & ${T}$ & category \\
\hline
ify & ke:keq, qemayuq & \texttt{\$sisit\$} & \textbf{no overlap} \\
ind & cewek, burung & \texttt{\$burung}, \texttt{terbang\$} & \textbf{overlap} \\
akb & unggas & \texttt{\$unggas\$} & \textbf{match} \\
afr & voël, vliegtuig & \texttt{voël} & \textbf{match} \\
\hline
\end{tabular}
\caption{\label{tab:panlex_examples}
Examples of languages that are classified into categories \textbf{no overlap}, \textbf{overlap} and \textbf{match} respectively for PanLex annotation.
}
\end{table}

\begin{table}
\centering
\footnotesize
\setlength\tabcolsep{3pt}
\begin{tabular}{llll}
\hline
 $l$ & ${P}$ & ${T}$ & factor \\
\hline
uig & qush & \texttt{\foreignlanguage{russian}{қуш}} (qush) & \textbf{script} \\
mua & \textipa{\v{Z}ù:} & \texttt{juu} & \textbf{transcription} \\
lip & \textipa{OklObE} & \texttt{\textipa{baklObE}} & \textbf{prefix} \\
mse & layra & \texttt{layagi} & \textbf{suffix}\\
sbl & ma`nok & \texttt{manokmanok} & \textbf{reduplication}\\
\hline
\end{tabular}
\caption{\label{tab:panlex_not_exact_match}
Example of PanLex translations that have different forms caused by non-lexical factors
}
\end{table}

\subsubsection{Translations compared with PanLex}\seclabel{bird_second_eva}
For each language, the linguist annotator also checks the translation of ``bird'' provided by PanLex \citep{kamholz-etal-2014-panlex}\footnote{https://translate.panlex.org/}. The translations are available in 1,137 languages out of the 1,331 languages in PBC where we found translations of the focal concept \concept{bird}. 
We define the following four categories for the PanLex evaluation where
$P$ are the tranlations from PanLex and $T$ are
the FP hits (target-language strings).

\begin{itemize}
\setlength\itemsep{0.0em}

\item \textbf{no translation}: $P =\emptyset$, i.e., PanLex gives no translation.

\item \textbf{no overlap}: $T \cap P=\emptyset$, i.e., none of the FP hits is found in PanLex translations.

\item \textbf{overlap}: $0 < |T  \cap P| < |T|$, i.e., some but not all of the FP hits are found in PanLex translations.

\item \textbf{match}: $|T  \cap P| = |T|$, i.e., all the FP hits can be found in PanLex translations. Note that we do not require all the translations in PanLex to be present in our set of target strings, since PanLex often gives a very long list of translations and our goal is to use PanLex translations to confirm the strings we identified.
\end{itemize}
We show examples for each category (except for \textbf{no translation})  in Table \ref{tab:panlex_examples}. 
When deciding whether a translation from PanLex matches an FP hit, the linguist annotator does not only look for an exact match of strings but also takes the differences in scripts, transcriptions, and morphological forms into consideration. 
For languages with multiple writing systems, words are naturally transliterated into a unified script for them to be comparable. It has also been observed, that the same word can sometimes be transcribed differently in different sources, especially for low-resource languages that have no standard writing system. 
Therefore, ``\textipa{\v{Z}ù:}'' in PanLex and ``juu'' in our FP hit will still be considered as a match.
Furthermore, it is also possible that the PanLex translation uses a morphologically different form compared to our FP hit, such as dictionary form versus inflected form. Possible morphological processes such as affixation, reduplication, vowel mutation, vowel ellipsis, and metathesis have been taken into consideration.
We show some examples of these factors in Table \ref{tab:panlex_not_exact_match}. 

With careful examination, if the linguist annotator concludes that \methodname actually has found the same lexeme as the PanLex translation, and the difference between PanLex and our FP hit is merely attributed to the above-mentioned factors, they will still be considered as matches.

\begin{table}
\centering
\footnotesize
\setlength\tabcolsep{3pt}
\begin{tabular}{lccc}
\hline
model & Coverage(g) & Coverage(a) & trans. per $l$\\
\hline
\methodname & .93 & .95 & 1.6 \\
Eflomal 0 & .94 & .96 & 7.0 \\
Eflomal 1 & .81 & .82 & 2.9 \\
Eflomal 0.1 & .70 & .79 & 2.1 \\
\hline
\end{tabular}
\caption{\label{tab:bird_results_vs_eflomal}
Coverage \& the number of translations proposed per language on average of \methodname and eflomal \citep{ostling2016efficient} on `bird'. Coverage (g) (resp. (a)) denotes the coverage computed globally for all languages (resp. computed for each language separately and then averaged over all languages). Elfomal 1 (resp. 0.1) denotes filtering the translations proposed whose frequency <= 1 (resp. <= 10\% of the number of verses containing `bird').
}
\end{table}

\subsubsection{Translations compared with Eflomal}\seclabel{bird_third_eva}

We also compare the coverage and the average number of translations proposed per language of \methodname and eflomal \citep{ostling2016efficient}, a statistical word alignment tool, in Table \ref{tab:bird_results_vs_eflomal}. We collect words that are aligned with one of the strings representing `bird' in all verses where `bird' occurs for eflomal baselines.
The coverage means the fraction of verses containing `bird' covered by the set of proposed translations. Global coverage denotes that we compute the coverage directly in all verses regardless of language while average coverage denotes that we first compute the coverage for each language and then average over all languages. 
We notice that,
eflomal, without filtering any translations, obtains the highest global and average coverage, which can be regarded as the upper bound. However, the number of translations per language on average is so high: 7.0. After filtering some proposed translations by their frequencies (1 and 0.1), we observe a sudden drop in the coverage. This indicates that (1) eflomal can propose many wrong alignments and (2) some correct alignments have very small frequencies. Because of its word-level alignment nature, eflomal cannot take the possible morphological changes of the words into consideration at all. On the contrary, \methodname only proposes 1.6 translations on average while keeping the coverage very close to the upper bound, suggesting that \methodname can identify the strings (ngrams) that are most associated with the concept and alleviate the possible problems caused by, e.g., morphological changes, in many languages.

\subsection{Swadesh concepts}\seclabel{swadesh_evaluation}
We resort to NoRaRe \citep{tjuka2022linking}\footnote{https://norare.clld.org/} to find the available translations of the 32 Swadesh concepts in 39 languages (NoRaRe covers). For each concept and each language, we store a triple indexed by the concept-language pair:
$$<\text{concept}, \text{language}, \text{translation(s)}>$$
We finally obtain 582 triples for evaluation (NoRaRe does not provide translations of a concept in all covered languages).
We use
$T$ to represent the set of translations proposed
by \methodname and  $N$ for NoRaRe translations (as ground-truth translations) in the triple. When judging whether a translation in $N$ matches a translation in $T$ generated by \methodname, we do the match leniently to allow for morphological changes. Specifically, if a translation in $N$ is a substring of a translation in $T$ generated by \methodname or the other way around, we regard it as a successful match. This is because $N$ often provide the dictionary forms of the nouns but $T$ are generated automatically based on the actual Bible verses where the nouns can change their suffixes or prefixes quite often depending on their roles in the verses.

We are especially interested in the triples in which our identified strings $T$ do not match the ground-truth translations $R$, i.e., $T \cap R = \emptyset$. We sampled 10 such triples (we provide $N$ and $T$ for each triple) in Table \ref{tab:swadesh_errors}. We notice that there are cases where the ground-truth translations $N$ use different versions of transliterations. For example, \texttt{\$andjing} vs. ``anjing'', \texttt{\$daoen} vs. ``daun'' and \texttt{\$boelan} vs. ``bulan'' in Malay (msa). Moreover, there can be multiple equivalent translations for a concept, but $N$ just lists one of them which is not used (or not identified) in PBC, e.g., ``\begin{CJK*}{UTF8}{gbsn}颈\end{CJK*}'' is a simpler but more formal translation of `neck' but the $N$ only lists ``\begin{CJK*}{UTF8}{gbsn}脖子\end{CJK*}'' in Chinese (zho); the concept `path' can also be translated to ``rout'' and ``sentier'' in French (fra) but only ``chemin'' is given by the $N$. Therefore, we see that this evaluation compared with NoRaRe can actually underestimate the performance of our method. 

\begin{table}
\centering
\footnotesize
\begin{tabular}{llll}
\hline
concept & $l$ & $N$ & $T$ \\
\hline
`dog' & msa & anjing & \texttt{\$andjing} \\
`seed' & cym & hedyn & \texttt{\$had\$}, \texttt{\$heu} \\
`leaf' & msa & daun & \texttt{\$daoen} \\
`horn' & est & ruupor & \texttt{sarve} \\
`mouth' & tur & ağiz & \texttt{\$ağzı}, \texttt{\$ağız} \\
`neck' & zho & \begin{CJK*}{UTF8}{gbsn}脖子\end{CJK*} & \begin{CJK*}{UTF8}{gbsn}颈\end{CJK*} \\
`moon' & msa & bulan & \texttt{\$boelan} \\
`water' & cym & dŵr & \texttt{dwfr\$}, \texttt{dyfr} \\
`rain' & msa & hujan & \texttt{\$hoedjan}  \\
`path' & fra & chemin & \texttt{\$la\$rout}, \texttt{\$sentier} \\
\hline
\end{tabular}
\caption{\label{tab:swadesh_errors}
10 randomly selected examples of triples for which we obtain a score of 0 in the ``partial'' setting in Table \ref{tab:swadesh_results}.
}
\end{table}

\begin{table}
\setlength\tabcolsep{4pt}
\centering
\footnotesize
\begin{tabular}{lll}
\hline
Concepts & Concreteness & Stability\\
\hline
`fish' & 5.0 & 0.86\\
`bird' & 5.0 & 0.68\\
`dog' & 4.85 & 0.85\\
`tree' & 5.0 & 0.64\\
`seed' & 4.71 & 0.38\\
`leaf' & 5.0 & 0.74\\
`root' & 4.34 & 0.78\\
`flesh' & 4.59 & 0.36\\
`blood' & 4.86 & 0.69\\
`horn' & 5.0 & 0.82\\
`hair' & 4.97 & 0.77\\
`ear' & 5.0 & 0.46\\
`mouth' & 4.74 & 0.49\\
`tooth' & 4.89 & 0.91\\
`tongue' & 4.93 & 0.61\\
`foot' & 4.9 & 0.7\\
`knee' & 5.0 & 0.38\\
`belly' & 4.8 & 0.4\\
`neck' & 5.0 & 0.72\\
`breast' & 4.89 & 0.65\\
`sun' & 4.83 & 0.49\\
`moon' & 4.9 & 0.48\\
`star' & 4.69 & 0.87\\
`water' & 5.0 & 0.48\\
`rain' & 4.97 & 0.68\\
`stone' & 4.72 & 0.71\\
`cloud' & 4.54 & 0.68\\
`smoke' & 4.96 & 0.57\\
`path' & 4.41 & 0.35\\
`mountain' & 4.96 & 0.64\\
`white' & 3.89 & 0.77\\
`night' & 4.52 & 0.68\\
\hline
\end{tabular}
\caption{\label{tab:concept_stability1}
The concreteness and stability measure of our considered focal concepts: Swadesh32. Concreteness is from \citep{brysbaert2014concreteness} while stability is computed using the statistics obtained by \methodname. If the concreteness is NA, it means the concept is not included in the resource.
}
\end{table}

\begin{table}
\setlength\tabcolsep{4pt}
\centering
\footnotesize
\begin{tabular}{lll}
\hline
Concepts & Concreteness & Stability \\
\hline
`babe' & 3.67 & 0.59\\
`hypocrit' & 2.43 & 0.81\\
`soldier' & 4.72 & 0.49\\
`scroll' & 4.11 & 0.57\\
`demon' & 3.32 & 0.45\\
`boat' & 4.93 & 0.71\\
`olive' & 4.9 & 0.8\\
`prayer' & 3.28 & 0.32\\
`mercy' & 1.57 & 0.29\\
`trumpet' & 4.86 & 0.83\\
`angel' & 3.82 & 0.88\\
`prison' & 4.68 & 0.62\\
`savior' & 3.04 & 0.49\\
`tomb' & 4.73 & 0.61\\
`husband' & 4.11 & 0.47\\
`bride' & 4.63 & 0.69\\
`talent' & 2.19 & 0.83\\
`peace' & 1.62 & 0.72\\
`secret' & 2.19 & 0.57\\
`faith' & 1.63 & 0.59\\
`woe' & 1.96 & 0.8\\
`throne' & 4.64 & 0.62\\
`wisdom' & 1.53 & 0.54\\
`disciple' & 3.29 & 0.73\\
`obeisance' & NA & 0.37\\
`truth' & 1.96 & 0.4\\
`memor' & 2.83 & 0.53\\
`governor' & 4.07 & 0.52\\
`poor' & 2.7 & 0.63\\
`blind' & 4.03 & 0.77\\
`spiritual' & 1.79 & 0.33\\
`justice' & 1.45 & 0.34\\
`courage' & 1.52 & 0.53\\
`purpose' & 1.52 & 0.3\\
`generation' & 1.96 & 0.56\\
`contrary' & 1.56 & 0.46\\
`prophesy' & 2.11 & 0.41\\
`decision' & 2.19 & 0.36\\
`request' & 2.59 & 0.32\\
`weakness' & 2.59 & 0.55\\
`journey' & 2.57 & 0.39\\
`public' & 2.57 & 0.23\\
`appearance' & 2.57 & 0.55\\
`expression' & 2.54 & 0.51\\
`marriage' & 2.51 & 0.51\\
`wrath' & 2.42 & 0.4\\
`trouble' & 2.25 & 0.45\\
`promise' & 2.09 & 0.46\\
`power' & 2.04 & 0.41\\
`pleasure' & 2.04 & 0.35\\
`thought' & 1.97 & 0.39\\
\hline
\end{tabular}
\caption{\label{tab:concept_stability2}
The concreteness and stability measure of our considered focal concepts: Bible51. Concreteness is from \citep{brysbaert2014concreteness} while stability is computed using the statistics obtained by \methodname. If the concreteness is NA, it means the concept is not included in the resource.
}
\end{table}

\section{Infrastructure \& environment}
We ran all our computational experiments on a CPU server with 48 cores and 1024 GB of memory.
We used Python 3.6\footnote{https://www.python.org/} throughout our implementation of \methodname and for visualizations. Specifically, for fundamental scientific computing (e.g., computing $\chi^2$ scores), we used NumPy\footnote{https://numpy.org/}, SciPy\footnote{https://scipy.org/} and scikit-learn\footnote{https://scikit-learn.org/stable/} packages. For visualization, we used NetworkX\footnote{https://networkx.org/}(mainly for the crosslingual semantic fields) and Matplotlib\footnote{https://matplotlib.org/} packages.

\pagebreak

\section{Crosslingual semantic fields}
\seclabel{crosslingualsemanticfields}

\begin{figure*}
  \centering
  \includegraphics[width=1\textwidth]{figs/semantic_field1.pdf}
  \caption{Visualization of semantic field (1).}\label{fig:semanticfield1}
\end{figure*}

\begin{figure*}
  \centering
  \includegraphics[width=1\textwidth]{figs/semantic_field2.pdf}
  \caption{Visualization of semantic field (2).}\label{fig:semanticfield2}
\end{figure*}

\begin{figure*}
  \centering
  \includegraphics[width=1\textwidth]{figs/semantic_field3.pdf}
  \caption{Visualization of semantic field (3).}\label{fig:semanticfield3}
\end{figure*}

\begin{figure*}
  \centering
  \includegraphics[width=1\textwidth]{figs/semantic_field4.pdf}
  \caption{Visualization of semantic field (4).}\label{fig:semanticfield4}
\end{figure*}

\begin{figure*}
  \centering
  \includegraphics[width=1\textwidth]{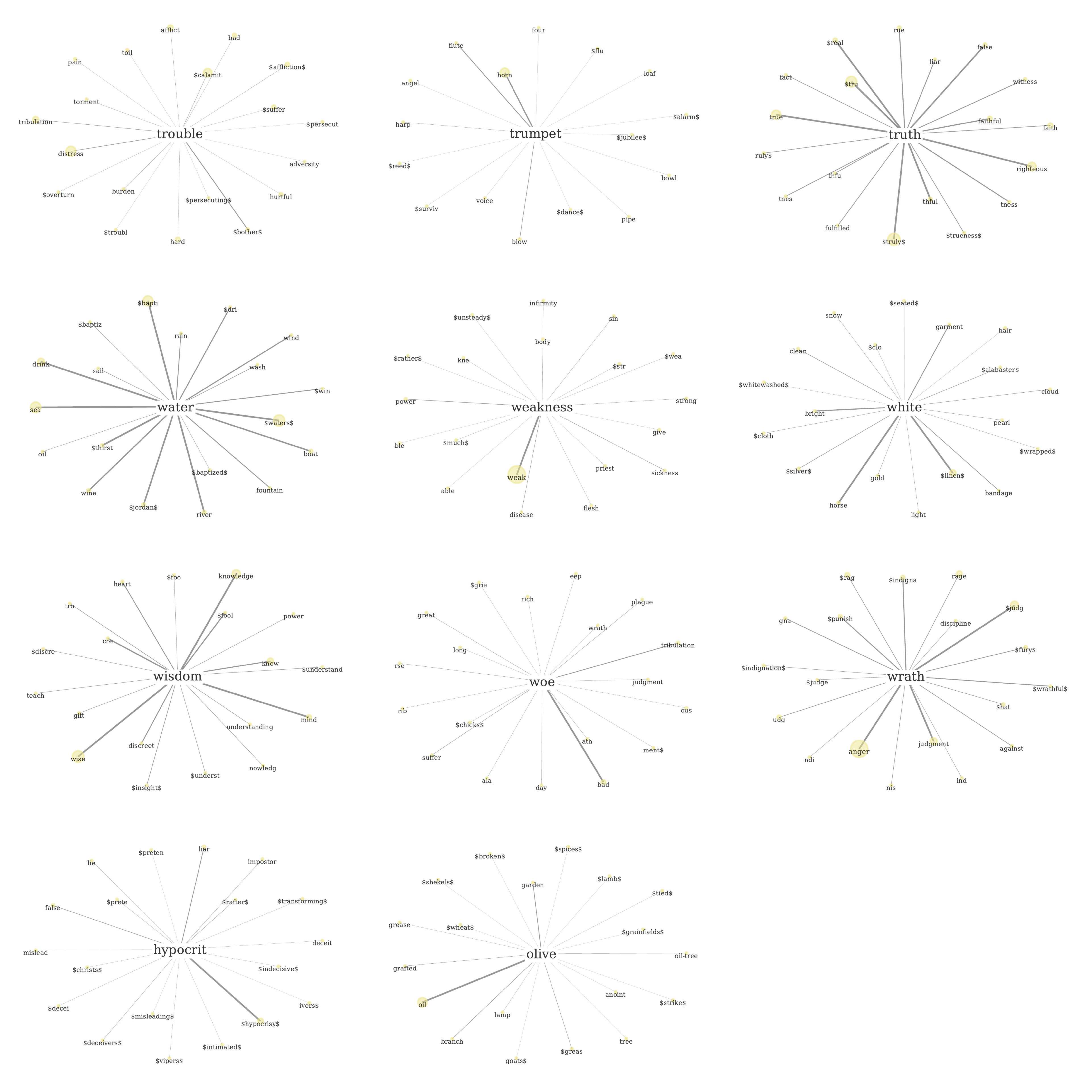}
  \caption{Visualization of semantic field (5).}\label{fig:semanticfield5}
\end{figure*}

\pagebreak
\section{Further analysis regarding language similarity}
\seclabel{furtheranalysislangsim}

\begin{figure*}[htbp]
\centering
\subfigure[Bible51 family]{
\begin{minipage}[t]{0.48\linewidth}
\centering
\includegraphics[width=3in]{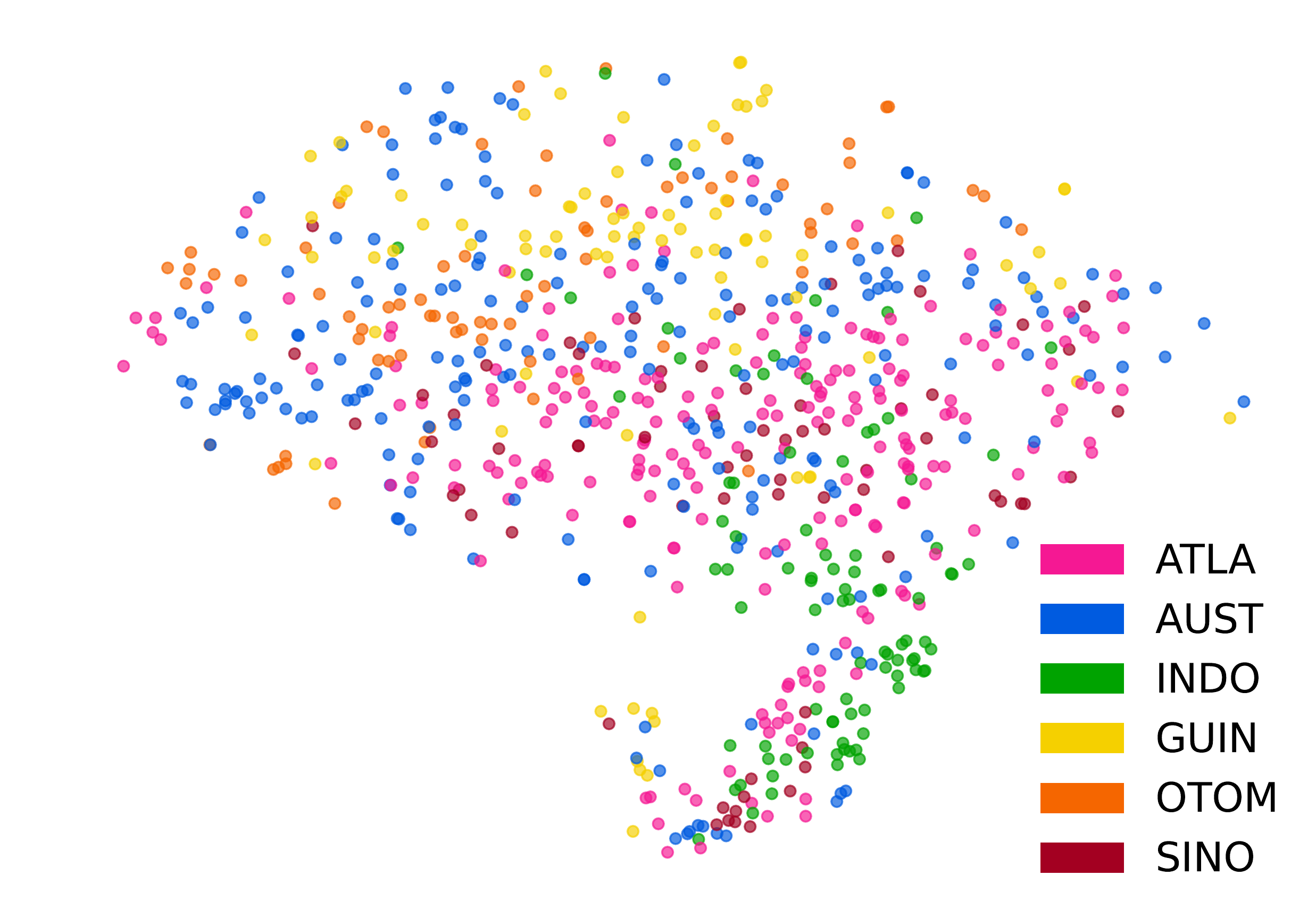}
\end{minipage}%
}%
\subfigure[All83 family]{
\begin{minipage}[t]{0.48\linewidth}
\centering
\includegraphics[width=3in]{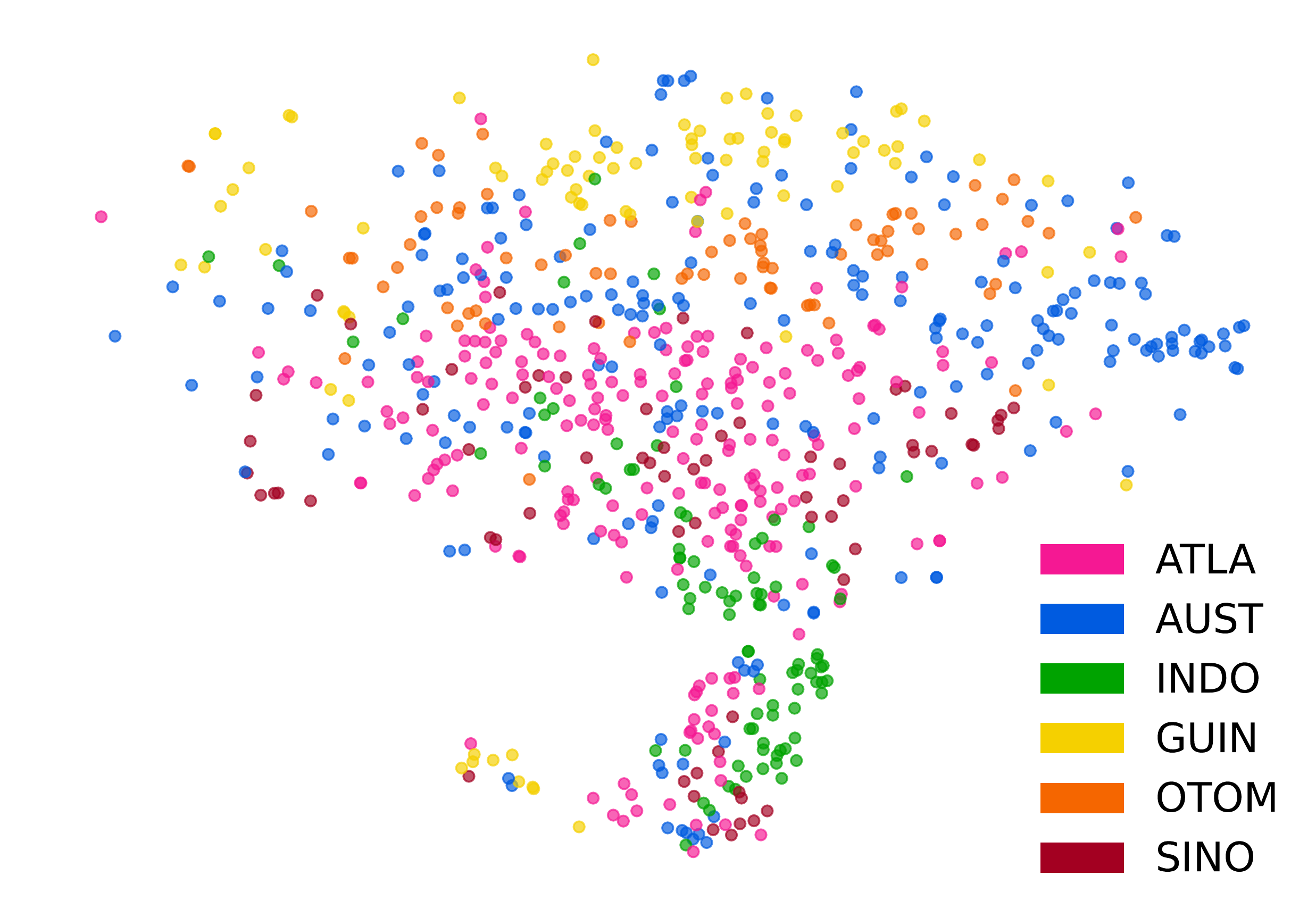}
\end{minipage}%
}%

\subfigure[Bible51 area]{
\begin{minipage}[t]{0.48\linewidth}
\centering
\includegraphics[width=3in]{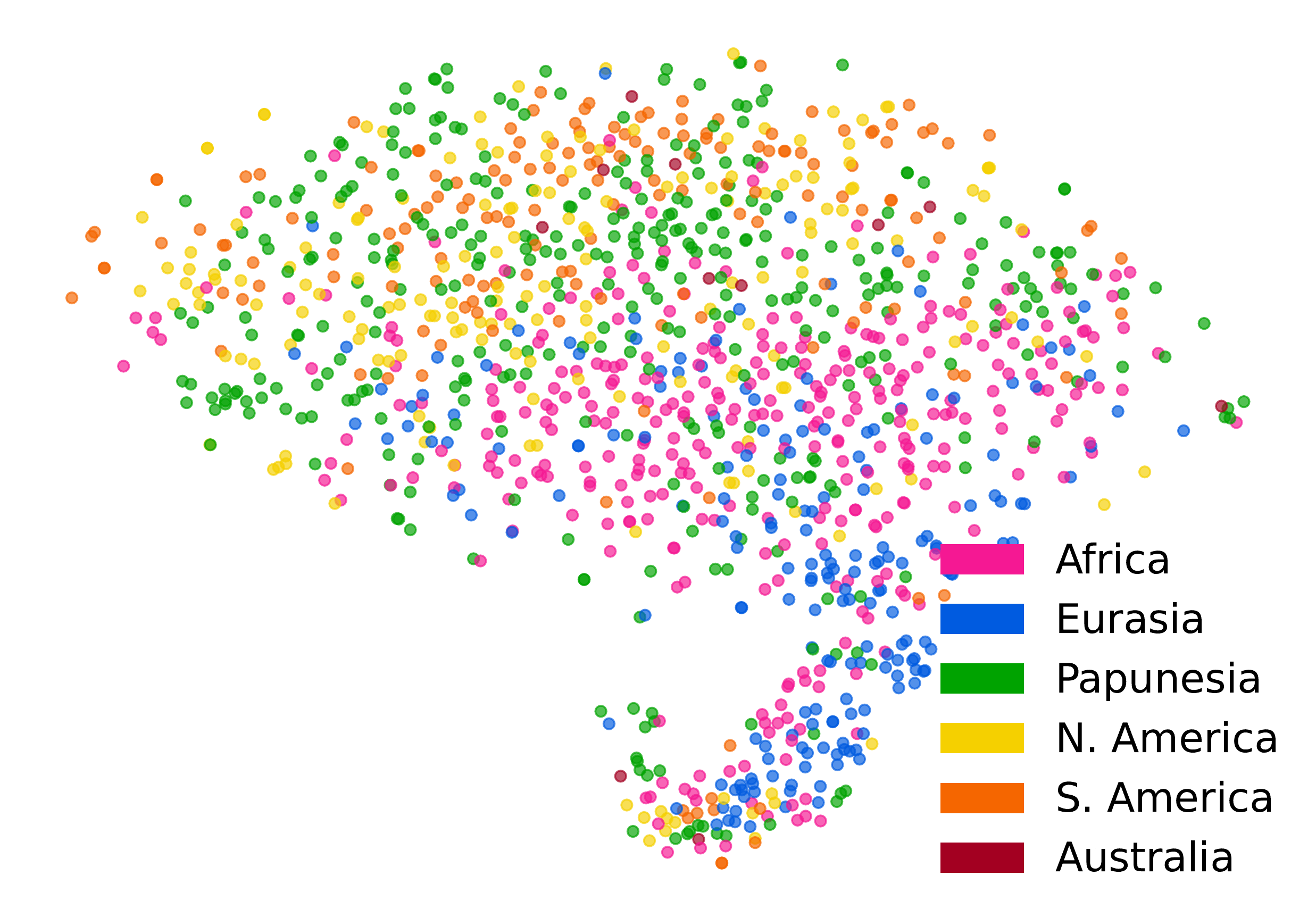}
\end{minipage}%
}%
\subfigure[All83 area]{
\begin{minipage}[t]{0.48\linewidth}
\centering
\includegraphics[width=3in]{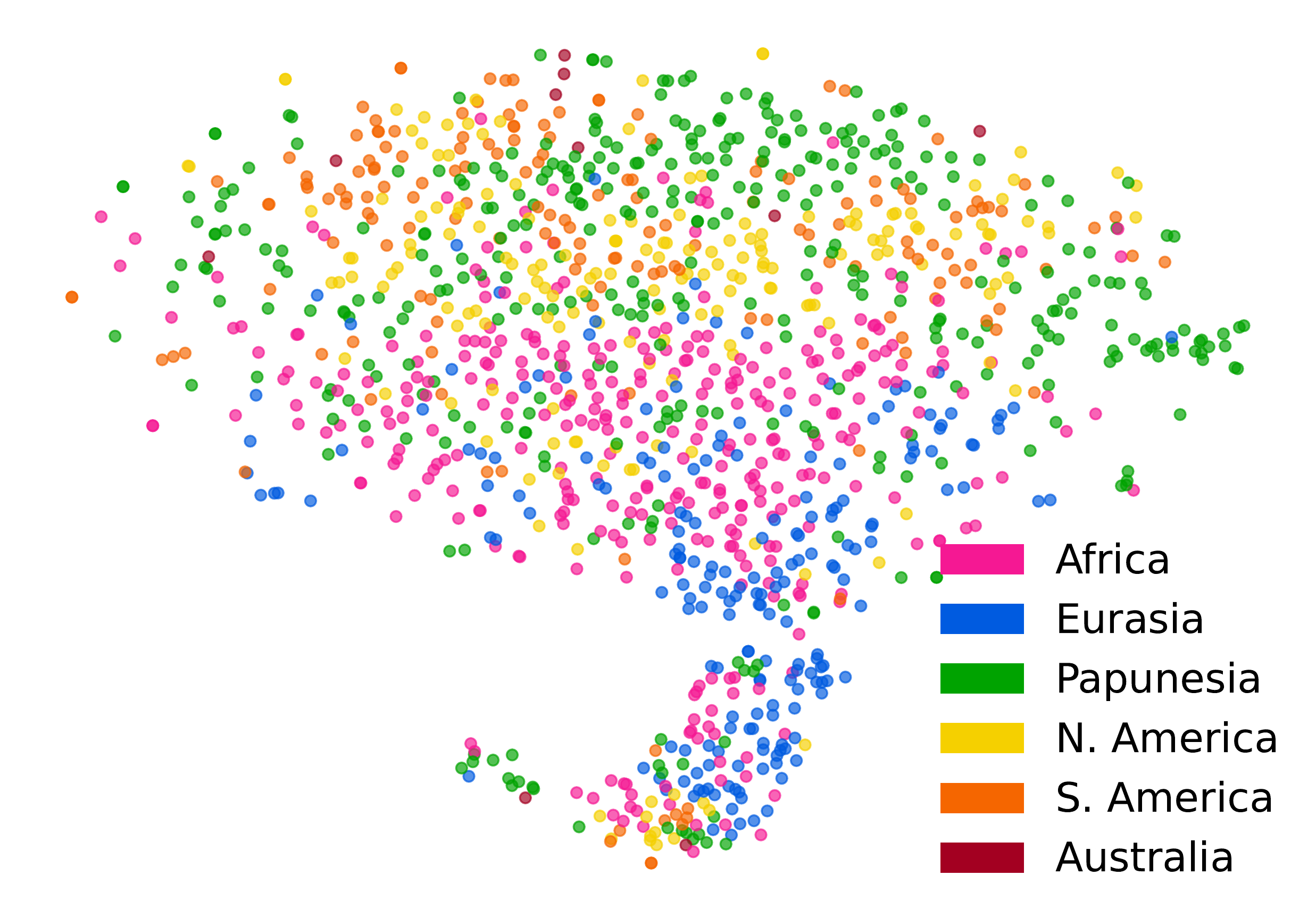}
\end{minipage}%
}%
\caption{t-SNE plot of the languages using \textbf{Swadesh32} and \textbf{All83} concatenation of concepts. The colors indicate different language families in the upper subfigure and different areas in the lower subfigure.}
\label{fig:clustering_51_83}
\end{figure*}

\begin{table*}
\centering
\setlength\tabcolsep{3pt}
\begin{tabular}{lcccccccc}
\hline
setting & global & atla & aust & indo & guin & otom & sino\\
\hline
Swadesh32 & 1280 & 222 & 215 & 91 & 87 & 76 & 68 \\
Bible51 & 1264 & 219 & 210 & 90 & 85 & 76 & 67 \\
All83  & 1263 & 219 & 210 & 90 & 85 & 76 & 67 \\
\hline
\end{tabular}
\caption{\label{tab:language_counts_by_setting}
Numbers of languages in the 6 families with over 50 languages. The numbers vary across the 3 concatenation settings as we only use languages for which all the considered concepts are available. Thus, \textbf{Swadesh32} has the most languages, and the number of languages decreases as we increase the number of concepts. We find that while both the \textcolor{aust_color}{Austronesian} and \textcolor{sino_color}{Sino-Tibetan} families are spread out, the binary classification result of \textcolor{aust_color}{Austronesian} languages is much better since the large geographical span is compensated by its higher number of languages.
}
\end{table*}

\begin{table*}
\centering
\setlength\tabcolsep{3pt}
\begin{tabular}{lcccccccc}
\hline
\#neighbors & global & Papunesia & Africa & Eurasia & North America & South America & Australia\\
\hline
k=1 & 0.51  & 0.51  & 0.52  & 0.64  & 0.52  & 0.35  & 0.00 \\
k=2 & 0.28  & 0.27  & 0.31  & 0.47  & 0.27  & 0.09  & 0.00 \\
k=3 & 0.45  & 0.44  & 0.51  & 0.62  & 0.43  & 0.22  & 0.00 \\
k=4 & 0.53  & 0.52  & 0.60  & 0.69  & 0.51  & 0.25  & 0.00 \\
k=5 & 0.55  & 0.54  & 0.61  & 0.72  & 0.57  & 0.28  & 0.09 \\
k=6 & 0.56  & 0.55  & 0.62  & 0.75  & 0.57  & 0.26  & 0.09 \\
k=7 & 0.56  & 0.55  & 0.63  & 0.74  & 0.56  & 0.24  & 0.09 \\
k=8 & 0.57  & 0.56  & 0.64  & 0.76  & 0.56  & 0.27  & 0.00 \\
k=9 & 0.57  & 0.54  & 0.66  & 0.77  & 0.59  & 0.26  & 0.00 \\
k=10 & 0.58  & 0.55  & 0.65  & 0.79  & 0.62  & 0.22  & 0.00 \\
\hline
\end{tabular}
\caption{\label{tab:area_classification1}
The result of binary classification (area) results using different numbers of nearest neighbors (1 to 10) of \textbf{Swadesh32}. The global column shows the results considering all languages. The rest columns denote the results only considering languages in that region.
}
\end{table*}

\begin{table*}
\centering
\setlength\tabcolsep{3pt}
\begin{tabular}{lcccccccc}
\hline
\#neighbors & global & Papunesia & Africa & Eurasia & North America & South America & Australia\\
\hline
k=1 & 0.49  & 0.54  & 0.57  & 0.63  & 0.28  & 0.30  & 0.00 \\
k=2 & 0.24  & 0.25  & 0.33  & 0.42  & 0.06  & 0.08  & 0.00 \\
k=3 & 0.41  & 0.43  & 0.59  & 0.62  & 0.13  & 0.09  & 0.00 \\
k=4 & 0.48  & 0.52  & 0.66  & 0.70  & 0.20  & 0.14  & 0.00 \\
k=5 & 0.51  & 0.55  & 0.70  & 0.72  & 0.18  & 0.18  & 0.00 \\
k=6 & 0.50  & 0.53  & 0.72  & 0.70  & 0.18  & 0.15  & 0.00 \\
k=7 & 0.50  & 0.53  & 0.71  & 0.70  & 0.18  & 0.16  & 0.00 \\
k=8 & 0.49  & 0.51  & 0.71  & 0.69  & 0.14  & 0.15  & 0.00 \\
k=9 & 0.48  & 0.48  & 0.73  & 0.71  & 0.12  & 0.12  & 0.00 \\
k=10 & 0.49  & 0.50  & 0.74  & 0.71  & 0.14  & 0.12  & 0.00 \\
\hline
\end{tabular}
\caption{\label{tab:area_classification2}
The result of binary classification (area) results using different numbers of nearest neighbors (1 to 10) of \textbf{Bible51}. The global column shows the results considering all languages. The rest columns denote the results only considering languages in that region.
}
\end{table*}

\begin{table*}
\centering
\setlength\tabcolsep{3pt}
\begin{tabular}{lcccccccc}
\hline
\#neighbors & global & Papunesia & Africa & Eurasia & North America & South America & Australia\\
\hline
k=1 & 0.54  & 0.59  & 0.65  & 0.68  & 0.38  & 0.25  & 0.00 \\
k=2 & 0.33  & 0.35  & 0.41  & 0.53  & 0.16  & 0.09  & 0.00 \\
k=3 & 0.51  & 0.51  & 0.69  & 0.71  & 0.27  & 0.17  & 0.00 \\
k=4 & 0.56  & 0.57  & 0.75  & 0.75  & 0.32  & 0.21  & 0.00 \\
k=5 & 0.58  & 0.59  & 0.77  & 0.78  & 0.34  & 0.22  & 0.00 \\
k=6 & 0.58  & 0.60  & 0.77  & 0.79  & 0.35  & 0.17  & 0.00 \\
k=7 & 0.58  & 0.60  & 0.79  & 0.81  & 0.34  & 0.15  & 0.00 \\
k=8 & 0.57  & 0.59  & 0.77  & 0.81  & 0.33  & 0.16  & 0.00 \\
k=9 & 0.58  & 0.59  & 0.78  & 0.81  & 0.32  & 0.18  & 0.00 \\
k=10 & 0.58  & 0.60  & 0.79  & 0.80  & 0.32  & 0.17  & 0.00 \\
\hline
\end{tabular}
\caption{\label{tab:area_classification3}
The result of binary classification (area) results using different numbers of nearest neighbors (1 to 10) of \textbf{All83}. The global column shows the results considering all languages. The rest columns denote the results only considering languages in that region.
}
\end{table*}

\begin{table*}
\centering
\setlength\tabcolsep{3pt}
\begin{tabular}{ccl}
\hline
\multicolumn{1}{c}{concept} & \multicolumn{1}{c}{lang.} & \multicolumn{1}{c}{ngrams} \\
\hline
\multirow{2}{*}{\concept{mouth}} & zho & \texttt{\$mouth\$}, \texttt{\$mouths\$}, \texttt{\$entrance}, 
                                \texttt{alat}, \texttt{\$ford} \\
                                & bod & \texttt{\$mouth\$}, \texttt{\$mouths\$}, \texttt{\$prais} \\ 
                                [1ex]
\multirow{2}{*}{\concept{neck}} & zho & \texttt{\$neck\$}, \texttt{\$necks\$}, \texttt{\$stiff-necked} \\
                                & bod & \texttt{\$neck\$}, \texttt{\$necks\$}, \texttt{\$obstina}, \texttt{\$stubbornness\$} \\
                                [1ex]
\multirow{2}{*}{\concept{tree}} & zho & \texttt{\$tree\$}, \texttt{\$trees\$}, \texttt{\$vine\$},
                                \texttt{-tree\$}, \texttt{\$boughs\$} \\
                                & bod & \texttt{\$tree\$}, \texttt{\$trees\$}, \texttt{\$chariot}, \texttt{wood}, \texttt{\$cedar}, \texttt{igs\$}, \texttt{\$timber} \\
                                [1ex]
\multirow{2}{*}{\concept{horn}} & zho & \texttt{\$horn\$}, \texttt{\$horns\$}, \texttt{\$corner} \\
                                & bod & \texttt{\$horn\$}, \texttt{\$horns\$}, \texttt{\$trumpet}, \texttt{\$fathering\$} \\
\hline
\end{tabular}
\caption{\label{tab:compare_sino}
Selected examples from the comparison of \textbf{Swadesh32} concepts of Mandarin Chinese (zho) and Tibetan (bod). Differences can be observed for several concepts, especially those related to body parts.
}
\end{table*}

\begin{table*}
\centering
\setlength\tabcolsep{3pt}
\begin{tabular}{ccl}
\hline
\multicolumn{1}{c}{concept} & \multicolumn{1}{c}{lang.} & \multicolumn{1}{c}{ngrams} \\
\hline
\multirow{5}{*}{\concept{fish}} & arb & \texttt{\$fish\$}, \texttt{\$fishes\$} \\
                               & pes & \texttt{\$fishes\$}, \texttt{\$fish\$}, \texttt{\$fish} \\
                               & tur & \texttt{\$fish\$}, \texttt{\$fishes\$}, \texttt{\$fish}, \texttt{\$honey} \\
                               & msa & \texttt{\$fishes\$}, \texttt{\$fish\$} \\
                               & ind & \texttt{\$fish\$}, \texttt{\$fishes\$} \\
                               [1ex]
\multirow{5}{*}{\concept{star}} & arb & \texttt{\$stars\$}, \texttt{\$star\$} \\
                               & pes & \texttt{\$stars\$}, \texttt{\$star\$} \\
                               & tur & \texttt{\$stars\$}, \texttt{\$star\$} \\
                               & msa & \texttt{\$stars\$}, \texttt{\$star\$} \\
                               & ind & \texttt{\$stars\$}, \texttt{\$star\$} \\
                               [1ex]
\multirow{5}{*}{\concept{blood}} & arb & \texttt{\$blood\$}, \texttt{\$bloodguilt\$} \\
                               & pes & \texttt{\$blood\$} \\
                               & tur & \texttt{\$blood\$}, \texttt{\$bloods}, \texttt{\$bloodguilt\$} \\
                               & msa & \texttt{\$blood\$} \\
                               & ind & \texttt{\$blood\$}, \texttt{\$blood} \\
                               [1ex]
\multirow{5}{*}{\concept{tongue}} & arb & \texttt{\$tongue\$}, \texttt{\$tongues\$} \\
                               & pes & \texttt{\$tongue\$}, \texttt{\$tongues\$} \\
                               & tur & \texttt{\$tongue\$}, \texttt{\$tongues\$}, \texttt{\$request}, \texttt{\$language} \\
                               & msa & \texttt{\$tongues\$}, \texttt{\$tongue\$}, \texttt{ebrew\$} \\
                               & ind & \texttt{\$tongue\$}, \texttt{\$tongues\$}, \texttt{\$language} \\
                               [1ex]
\multirow{5}{*}{\concept{bird}} & arb & \texttt{\$birds\$}, \texttt{\$bird\$}, \texttt{\$flying\$}, \texttt{\$fowls\$}                                 \\
                               & pes & \texttt{\$birds\$}, \texttt{\$bird\$} \\
                               & tur & \texttt{\$birds\$}, \texttt{\$bird\$}, \texttt{\$fowl} \\
                               & msa & \texttt{\$birds\$}, \texttt{\$bird\$}, \texttt{dove}, \texttt{\$sparrows\$}, \texttt{\$eagle} \\
                               & ind & \texttt{\$birds\$}, \texttt{\$bird\$}, \texttt{\$eagle}, \texttt{\$turtledove}, \texttt{\$ostrich}, \texttt{\$raven}, \texttt{\$bird} \\
\hline
\end{tabular}
\caption{\label{tab:compare_islamization}
Selected examples from the comparison of \textbf{Swadesh32} concepts of several languages influenced by Islam. arb: Standard Arabic, pes: Western Farsi, tur: Turkish, msa: Standard Malay, ind: Standard Indonesian.
}
\end{table*}

\begin{table*}
\centering
\setlength\tabcolsep{3pt}
\begin{tabular}{ccl}
\hline
\multicolumn{1}{c}{concept} & \multicolumn{1}{c}{lang.} & \multicolumn{1}{c}{ngrams} \\
\hline
\multirow{7}{*}{\concept{blood}} & eng & \texttt{\$blood\$} \\
                               & spa & \texttt{\$blood\$} \\
                               & ell & \texttt{\$blood\$} \\
                               & rus & \texttt{\$blood\$}, \texttt{\$blood} \\
                               & tgl & \texttt{\$blood\$} \\
                               & swh & \texttt{\$blood\$}, \texttt{\$blood} \\
                               & hye & \texttt{\$blood\$}, \texttt{\$blood} \\
                               [1ex]
\multirow{7}{*}{\concept{tongue}} & eng & \texttt{\$tongue\$}, \texttt{\$tongues\$} \\
                               & spa & \texttt{\$tongue\$}, \texttt{\$tongues\$}, \texttt{\$language} \\
                               & ell & \texttt{\$tongue\$}, \texttt{\$tongues\$}, \texttt{\$language} \\
                               & rus & \texttt{\$tongue\$}, \texttt{\$tongues\$}, \texttt{\$language} \\
                               & tgl & \texttt{\$tongue\$}, \texttt{\$tongues\$}, \texttt{\$language} \\
                               & swh & \texttt{\$tongue\$}, \texttt{\$tongues\$}, \texttt{\$language} \\
                               & hye & \texttt{\$tongue\$}, \texttt{\$tongues\$}, \texttt{\$language} \\
                               [1ex]
\multirow{7}{*}{\concept{bird}} & eng & \texttt{\$birds\$}, \texttt{\$bird\$} \\
                               & spa & \texttt{\$birds\$}, \texttt{\$bird\$}, \texttt{\$fowls\$} \\
                               & ell & \texttt{\$birds\$}, \texttt{\$bird\$}, \texttt{\$bird} \\
                               & rus & \texttt{\$birds\$}, \texttt{\$bird\$}, \texttt{\$fowl}, \texttt{\$bird} \\
                               & tgl & \texttt{\$birds\$}, \texttt{\$bird\$}, \texttt{\$fowl} \\
                               & swh & \texttt{\$birds\$}, \texttt{\$bird\$}, \texttt{\$fowl} \\
                               & hye & \texttt{\$birds\$}, \texttt{\$bird\$}, \texttt{\$flying\$}, \texttt{\$fowl} \\
\hline
\end{tabular}
\caption{\label{tab:compare_christianity}
Selected examples from the comparison of \textbf{Swadesh32} concepts of several languages influenced by Christianity. eng: English, spa: Spanish, ell: Modern Greek, rus: Russian, tgl: Tagalog, swh: Swahili, hye: Eastern Armenian.
}
\end{table*}

\begin{table*}
\centering
\setlength\tabcolsep{3pt}
\begin{tabular}{ccl}
\hline
\multicolumn{1}{c}{concept} & \multicolumn{1}{c}{lang.} & \multicolumn{1}{c}{ngrams} \\
\hline
\multirow{6}{*}{\concept{blood}} & eng & \texttt{\$blood\$} \\
                               & deu & \texttt{\$blood\$} \\
                               & fra & \texttt{\$blood\$} \\
                               & jpn & \texttt{\$blood\$}, \texttt{\$blood} \\
                               & kor & \texttt{\$blood\$}, \texttt{\$escap}, \texttt{\$skin\$}, \texttt{\$airs\$}, \texttt{\$pip}, \texttt{\$flut} \\
                               & zho & \texttt{\$blood\$}, \texttt{\$blood} \\
                               [1ex]
\multirow{6}{*}{\concept{tongue}} & eng & \texttt{\$tongue\$}, \texttt{\$tongues\$} \\
                               & deu & \texttt{\$tongue\$}, \texttt{\$tongues\$} \\
                               & fra & \texttt{\$tongue\$}, \texttt{\$tongues\$}, \texttt{\$language} \\
                               & jpn & \texttt{\$tongue\$}, \texttt{\$tongues\$} \\
                               & kor & \texttt{\$tongue\$}, \texttt{\$tongues\$}, \texttt{\$language} \\
                               & zho & \texttt{\$tongue\$}, \texttt{\$tongues\$}, \texttt{\$language} \\
                               [1ex]
\multirow{6}{*}{\concept{mouth}} & eng & \texttt{\$mouth\$}, \texttt{\$mouths\$} \\
                               & deu & \texttt{\$mouth\$}, \texttt{\$mouths\$} \\
                               & fra & \texttt{\$mouth\$}, \texttt{\$mouths\$} \\
                               & jpn & \texttt{\$mouth\$}, \texttt{\$mouths\$}, \texttt{\$entrance}, \texttt{\$kiss}, \texttt{\$whistl}, \texttt{\$doorkeeper}, \texttt{\$contention} \\
                               & kor & \texttt{\$mouth\$}, \texttt{\$mouths\$}, \texttt{\$entrance}, \texttt{\$lip}, \texttt{\$clothe}, \texttt{\$kiss}, \texttt{\$overla} \\
                               & zho & \texttt{\$mouth\$}, \texttt{\$mouths\$}, \texttt{\$entrance}, \texttt{alat}, \texttt{\$ford} \\
\hline
\end{tabular}
\caption{\label{tab:compare_asia}
Selected examples from the comparison of \textbf{Swadesh32} concepts of languages possibly influenced by western and Chinese languages. eng: English, deu: German, fra: French, jpn: Japanese, kor: Korean, zho: Mandarin Chinese.
}
\end{table*}

\begin{table*}
\centering
\setlength\tabcolsep{3pt}
\begin{tabular}{ccl}
\hline
\multicolumn{1}{c}{concept} & \multicolumn{1}{c}{lang.} & \multicolumn{1}{c}{ngrams} \\
\hline
\multirow{4}{*}{\concept{bird}} & spa & \texttt{\$birds\$}, \texttt{\$bird\$}, \texttt{\$fowls\$} \\
                               & tgl & \texttt{\$birds\$}, \texttt{\$bird\$}, \texttt{\$fowl} \\
                               & ceb & \texttt{\$birds\$}, \texttt{\$bird\$}, \texttt{\$fowl} \\
                               & hil & \texttt{\$birds\$}, \texttt{\$bird\$}, \texttt{\$sparrows\$} \\
                               [1ex]
\multirow{4}{*}{\concept{ear}} & spa & \texttt{\$ear\$}, \texttt{\$ears\$}, \texttt{\$heard\$} \\
                               & tgl & \texttt{\$ears\$}, \texttt{\$ear\$}, \texttt{\$listen\$}, \texttt{\$hearing\$} \\
                               & ceb & \texttt{\$ears\$}, \texttt{\$ear\$}, \texttt{\$hear\$}, \texttt{\$hearing\$} \\
                               & hil & \texttt{\$ear\$}, \texttt{\$ears\$}, \texttt{\$hear} \\
                               [1ex]
\multirow{4}{*}{\concept{tongue}} & spa & \texttt{\$tongue\$}, \texttt{\$tongues\$}, \texttt{\$language} \\
                               & tgl & \texttt{\$tongue\$}, \texttt{\$tongues\$}, \texttt{\$language} \\
                               & ceb & \texttt{\$tongue\$}, \texttt{\$tongues\$}, \texttt{\$language} \\
                               & hil & \texttt{\$tongue\$}, \texttt{\$tongues\$} \\
\hline
\end{tabular}
\caption{\label{tab:compare_philippine}
Selected examples from the comparison of \textbf{Swadesh32} concepts of several Philippine languages influenced by Spanish. spa: Spanish, tgl: Tagalog, ceb: Cebuano, hil: Hiligaynon.
}
\end{table*}

\begin{table*}
\centering
\setlength\tabcolsep{3pt}
\begin{tabular}{ccl}
\hline
\multicolumn{1}{c}{concept} & \multicolumn{1}{c}{lang.} & \multicolumn{1}{c}{ngrams} \\
\hline
\multirow{4}{*}{\concept{tree}} & yor & \texttt{\$tree\$}, \texttt{\$trees\$}, \texttt{wood}, \texttt{\$stake}, \texttt{\$frankincense\$}, \texttt{\$thornbush}, \texttt{\$palm-tree\$} \\
                               & ibo & \texttt{\$tree\$}, \texttt{\$trees\$}, \texttt{\$pole}, \texttt{wood}, \texttt{\$impal}, \texttt{\$stake}, \texttt{\$panel} \\
                               & mcn & \texttt{\$tree\$}, \texttt{\$trees\$}, \texttt{wood}, \texttt{\$stake}, \texttt{\$impale}, \texttt{\$cedar}, \texttt{\$timber} \\
                               & twi & \texttt{\$tree\$}, \texttt{\$trees\$}, \texttt{\$wood}, \texttt{\$panel\$}, \texttt{\$pole}, \texttt{\$figs\$}, \texttt{\$timber} \\
                               [1ex]
\multirow{4}{*}{\concept{hair}} & yor & \texttt{\$hair\$}, \texttt{\$hairs\$}, \texttt{\$wool\$} \\
                               & ibo & \texttt{\$hair\$}, \texttt{\$hairs\$}, \texttt{\$wool}, \texttt{\$shear}, \texttt{\$beard} \\
                               & mcn & \texttt{\$hair\$}, \texttt{\$hairs\$}, \texttt{\$wool\$}, \texttt{\$shave}, \texttt{\$baldness\$}, \texttt{\$shear}, \texttt{goat} \\
                               & twi & \texttt{\$hair\$}, \texttt{\$hairs\$}, \texttt{\$beard}, \texttt{\$shave}, \texttt{\$head\$}, \texttt{\$wool} \\
                               [1ex]
\multirow{4}{*}{\concept{mouth}} & yor & \texttt{\$mouth\$}, \texttt{\$mouths\$}, \texttt{\$entrance}, \texttt{\$kiss}, \texttt{\$palate\$}, \texttt{\$marvel}, \texttt{\$suckling} \\
                               & ibo & \texttt{\$mouth\$}, \texttt{\$mouths\$}, \texttt{\$gate}, \texttt{\$entrance}, \texttt{\$lip}, \texttt{curse}, \texttt{\$precious\$} \\
                               & mcn & \texttt{\$mouth\$}, \texttt{\$mouths\$}, \texttt{\$lips\$}, \texttt{fulfill}, \texttt{\$denie}, \texttt{\$disown}, \texttt{\$entrance} \\
                               & twi & \texttt{\$mouth\$}, \texttt{\$mouths\$}, \texttt{\$gat}, \texttt{\$collect}, \texttt{\$lip}, \texttt{\$entrance}, \texttt{\$registered\$} \\
\hline
\end{tabular}
\caption{\label{tab:compare_afro}
Selected examples from the comparison of \textbf{Swadesh32} concepts of four African languages. yor: Yoruba, ibo: Igbo, mcn: Masana, twi: Twi.
}
\end{table*}

\begin{table*}
\centering
\setlength\tabcolsep{3pt}
\begin{tabular}{ccl}
\hline
\multicolumn{1}{c}{concept} & \multicolumn{1}{c}{lang.} & \multicolumn{1}{c}{ngrams} \\
\hline
\multirow{3}{*}{\concept{bird}} & cak & \texttt{\$tree\$}, \texttt{\$trees\$}, \texttt{wood}, \texttt{\$pole}, \texttt{\$cedar}, \texttt{\$figs\$}, \texttt{\$palm-tree\$} \\
                               & kjb & \texttt{\$tree\$}, \texttt{\$trees\$}, \texttt{\$cedar}, \texttt{\$panel}, \texttt{\$wood}, \texttt{\$figs\$}, \texttt{\$pole} \\
                               & tzj & \texttt{\$tree\$}, \texttt{\$trees\$} \\
                               [1ex]
\multirow{3}{*}{\concept{seed}} & cak & \texttt{\$seed\$}, \texttt{\$seeds\$}, \texttt{braham}, \texttt{\$vine}, \texttt{\$sow}, \texttt{fruit}, \texttt{\$harvest} \\
                               & kjb & \texttt{\$seed\$}, \texttt{\$seeds\$}, \texttt{braham}, \texttt{\$garden}, \texttt{fruits\$}, \texttt{\$vine}, \texttt{\$harvest} \\
                               & tzj & \texttt{\$seed\$}, \texttt{\$seeds\$}, \texttt{\$sow}, \texttt{\$harvest\$}, \texttt{fruit} \\
                               [1ex]
\multirow{3}{*}{\concept{knee}} & cak & \texttt{\$knees\$}, \texttt{\$bow}, \texttt{\$worship}, \texttt{\$trembl}, \texttt{fell\$} \\
                               & kjb & \texttt{\$knees\$}, \texttt{\$knee\$}, \texttt{\$obeisance\$}, \texttt{\$worship}, \texttt{\$bow}, \texttt{\$fell\$}  \\
                               & tzj & \texttt{\$knees\$}, \texttt{\$worship}, \texttt{\$obeisance\$}, \texttt{\$fell\$} \\
\hline
\end{tabular}
\caption{\label{tab:compare_mayan}
Selected examples from the comparison of \textbf{Swadesh32} concepts of three Mayan languages. cak: Kaqchikel, kjb: Q'anjob'al, tzj: Tz'utujil.
}
\end{table*}

\begin{table*}
\centering
\setlength\tabcolsep{3pt}
\begin{tabular}{ccll}
\hline
\multicolumn{1}{c}{concept} & \multicolumn{1}{c}{target lang.} & \multicolumn{1}{c}{trans. in target lang. (eng)} \\
\hline
\multirow{4}{*}{\concept{mouth}} & jpn & \begin{CJK*}{UTF8}{bsmi}口\end{CJK*} (mouth, opening, entrance) \\
                               & kor &\begin{CJK*}{UTF8}{mj}구 (口)\end{CJK*} (entrance, gate, mouth) \\
                               & zho & \begin{CJK*}{UTF8}{bsmi}口 (mouth, gate, entrance), 嘴 (mouth, lips) \end{CJK*} \\
                               & fra & bouche (mouth) \\
                               [1ex]
\multirow{5}{*}{\concept{tongue}} & spa & lengua (tongue, language) \\
                               & tgl & dilà (tongue, language), wikà (tongue, language) \\
                               & ceb & dila (tongue), pinulongan (tongue, language) \\
                               & hil & dilà (tongue), dilâ (tongue) \\
                               & msa & lidah (tongue), oojoo leeda (tongue) \\
                               [1ex]
\hline
\end{tabular}
\caption{\label{tab:compare_word_correspondences}
We use online dictionaries such as PanLex and Google Translate to look up two example concepts in the target languages and verify the associated meanings of their translations in English. We show English translations considering all used sources. For example, we obtain different translations for Tagalog ``dilà'' and ``wikà'' depending on the dictionary source and find that they can possibly mean both ``tongue'' and ``language''. The target language translations are consistent with our findings: the three East Asian languages (jpn, kor, zho) share a common conceptualization of \texttt{mouth} as \texttt{entrance}, which is missing for French (fra); similar to Spanish (spa), some Philippine languages (tgl, ceb) conceptualize \texttt{tongue} as \texttt{language}, whereas another \textcolor{aust_color}{Austronesian} language, Standard Malay (msa), does not.
}
\end{table*}

\end{document}